\documentclass[11pt,letterpaper]{article}

\usepackage[margin=1in]{geometry}
\usepackage[T1]{fontenc}
\usepackage[utf8]{inputenc}
\usepackage{times}
\usepackage{microtype}
\usepackage[numbers,sort&compress]{natbib}
\usepackage{graphicx}
\usepackage{booktabs}
\usepackage{tabularx}
\usepackage{array}
\usepackage{multirow}
\usepackage{amsmath,amssymb}
\usepackage{xcolor}
\usepackage{enumitem}
\usepackage{seqsplit}
\usepackage{xurl}
\usepackage{caption}
\usepackage{subcaption}
\usepackage{longtable}
\usepackage{threeparttable}
\usepackage{float}
\usepackage[section]{placeins}
\usepackage{hyperref}
\newcommand{\textcite}[1]{\citet{#1}}
\hypersetup{colorlinks=true,linkcolor=blue!55!black,citecolor=blue!55!black,urlcolor=blue!55!black}

\newcommand{\Opl}{Overton Pluralism}
\newcommand{\Spl}{Steerable Pluralism}
\newcommand{\Dpl}{Distributional Pluralism}
\newcolumntype{L}[1]{>{\raggedright\arraybackslash}p{#1}}

\title{\textbf{Pluralistic Alignment in Large Language Models: Objectives, Human References, and Preference Representations}}
\author{
    \textbf{Meng-Chen Wu}\thanks{This work is unrelated to the authors' positions at Amazon.},
    \textbf{Umesh Bodhwani}\footnotemark[1] \\
     Amazon Science\\
 \texttt{\{mengchw, bodhwani\}@amazon.com}
    } 
\date{}

\begin{document}
\maketitle
\begin{abstract}
Pluralistic alignment aims to represent meaningful variation in human preferences rather than collapse heterogeneous judgments into a single target. Yet existing work often mixes three distinct questions: what pluralistic behavior a model should exhibit, whose preferences define the reference, and what evidence is sufficient to demonstrate success. We organize the literature around a two-layer taxonomy. Layer~1 distinguishes representing a range of human-supported viewpoints (\Opl), adapting faithfully to a specified perspective or preference condition (\Spl), and matching the response distribution observed in a target population (\Dpl). Layer~2 distinguishes aggregate populations, countries and regions, and subgroups as human references. Preference representation, implementation mechanism, modality, and evidence quality are treated as cross-cutting design choices rather than additional forms of pluralism. We synthesize a protocol-bounded corpus of 70 works on pluralistic alignment and neighboring areas, using the objective-by-human-reference matrix to map existing approaches and identify underexplored combinations. The coded literature concentrates on \Spl{} and \Dpl, while eligible \Opl{} work appears only at the aggregate-population level under our stated protocol. We then connect each objective to the evidence needed to support it, including perspective validity, conditional fidelity, distributional fidelity, support preservation, and uncertainty in subgroup estimates. Together, the taxonomy and evaluation framework clarify what current methods establish, where their claims exceed their evidence, and how future systems can represent disagreement without mistaking variation for faithful representation.
\end{abstract}

\section{Introduction}

Human preferences are diverse. People differ in their political priorities, moral intuitions, cultural expectations, and everyday judgments, and meaningful variation can persist even among individuals with similar demographic backgrounds. Language-model alignment, however, often aggregates these heterogeneous judgments into a single preferred response or reward signal. Such aggregation can be useful when a system must ultimately choose an action, but it collapses an important distinction: \emph{optimizing for one broadly acceptable response is not the same as representing the range and prevalence of preferences among the people a model is intended to serve.}

Pluralistic alignment addresses this mismatch by treating human disagreement as part of the alignment target rather than as noise to be averaged away. Yet pluralistic alignment is not a single objective. Prior work distinguishes three forms of pluralistic behavior \citep{sorensen2024roadmap}. \Opl{} seeks to represent the range of human-supported viewpoints on a contested issue. \Spl{} asks whether a model responds faithfully when conditioned on a particular perspective, group, role, or value. \Dpl{} aims to reproduce how frequently different responses occur in a reference population. These objectives are related but not interchangeable. A response can be diverse while including viewpoints unsupported by the relevant population; a model can react to a demographic prompt while relying on stereotypes rather than observed preferences; and a model can match an aggregate response distribution while systematically misrepresenting particular groups.

The objective alone, however, does not fully specify the alignment target. A pluralistic claim must also state \emph{whose preferences define success}. An aggregate population, a country or region, and a demographic subgroup correspond to different empirical quantities, even when they are derived from the same underlying survey. Aggregate estimates can conceal meaningful subgroup differences, while increasingly fine partitions can introduce sampling uncertainty and encourage unsupported assumptions about group members. Individual personalization raises a related but distinct question. Accurately predicting what a particular user prefers does not by itself show that a system represents disagreement in a broader population. Without an explicit human reference, it is difficult to determine what kind of pluralistic behavior a method establishes or whether its evaluation provides evidence for that claim.

We therefore organize pluralistic alignment along two complementary layers. \textbf{Layer~1} specifies the alignment objective: \Opl, \Spl, or \Dpl. \textbf{Layer~2} specifies the empirical human reference: aggregate populations, countries and regions, or subgroups. Crossing these dimensions yields an objective-by-human-reference matrix that makes pluralistic claims more precise. The first dimension specifies \emph{what the model should do with disagreement}; the second specifies \emph{whose judgments determine success}. Other design choices cut across these layers rather than defining additional forms of pluralism. Attributes, observed judgments, value profiles, and latent preference models determine how human preferences are represented, while prompting, parameter learning, and output composition determine how those representations influence model behavior. Figure~\ref{fig:framework} situates this two-layer taxonomy relative to neighboring work on perspectivist NLP, diversity-aware cultural alignment, and personalization.

This survey makes three contributions. \emph{Conceptually}, we separate pluralistic objectives from human references and use this distinction to clarify the relationship between population-level pluralism and individual personalization. \emph{Empirically}, we organize a protocol-bounded body of work using the objective-by-human-reference matrix, while separately accounting for value-, community-, persona-, and latent-preference approaches that fall outside the matrix's narrower eligibility criteria. \emph{Methodologically}, we connect each objective to the evidence required to substantiate it: perspective validity for \Opl, target fidelity for \Spl, and distributional fidelity together with support preservation for \Dpl. In this way, the survey complements prior reviews of user modeling and broader pluralistic capabilities by asking not only what existing methods do, but also \emph{which claims about human preferences their evidence can support} \citep{xie2025survey,shetty2026pluralsurvey}.

We review work publicly available through October 14, 2026. The protocol-bounded corpus contains 70 direct or adjacent works, of which 29 satisfy the stricter eligibility criteria for the objective-by-human-reference matrix, yielding 38 paper-to-cell assignments. These counts describe the corpus identified under our protocol rather than an exhaustive census of the field. The appendices document the search strategy, coding decisions, and sensitivity analyses. The remainder of the paper follows the same conceptual progression. Section~\ref{sec:framework} formalizes the alignment objectives and human references; Sections~\ref{sec:representation} and~\ref{sec:methods} examine how human variation is represented and how those representations enter a system; Section~\ref{sec:evaluation} develops the evidence needed to evaluate the resulting behavior; and Section~\ref{sec:landscape} uses the objective-by-human-reference matrix to characterize the literature and identify open directions.

\section{A Two-Layer Taxonomy of Pluralistic Alignment}\label{sec:framework}

Building on the distinction introduced above, we formalize pluralistic alignment along two complementary layers: the behavioral objective and the human reference against which that objective is evaluated. The first asks \emph{what the model should do with disagreement}; the second asks \emph{whose judgments define success}. Specifying both dimensions turns a broad appeal to ``pluralism'' into a testable claim. For example, country-conditioned \Dpl{} requires agreement with the observed response frequencies of a specified country, whereas subgroup-conditioned \Opl{} requires coverage of perspectives supported or validated by members of a specified subgroup.

\begin{figure}[!t]
\centering
\includegraphics[width=0.99\linewidth]{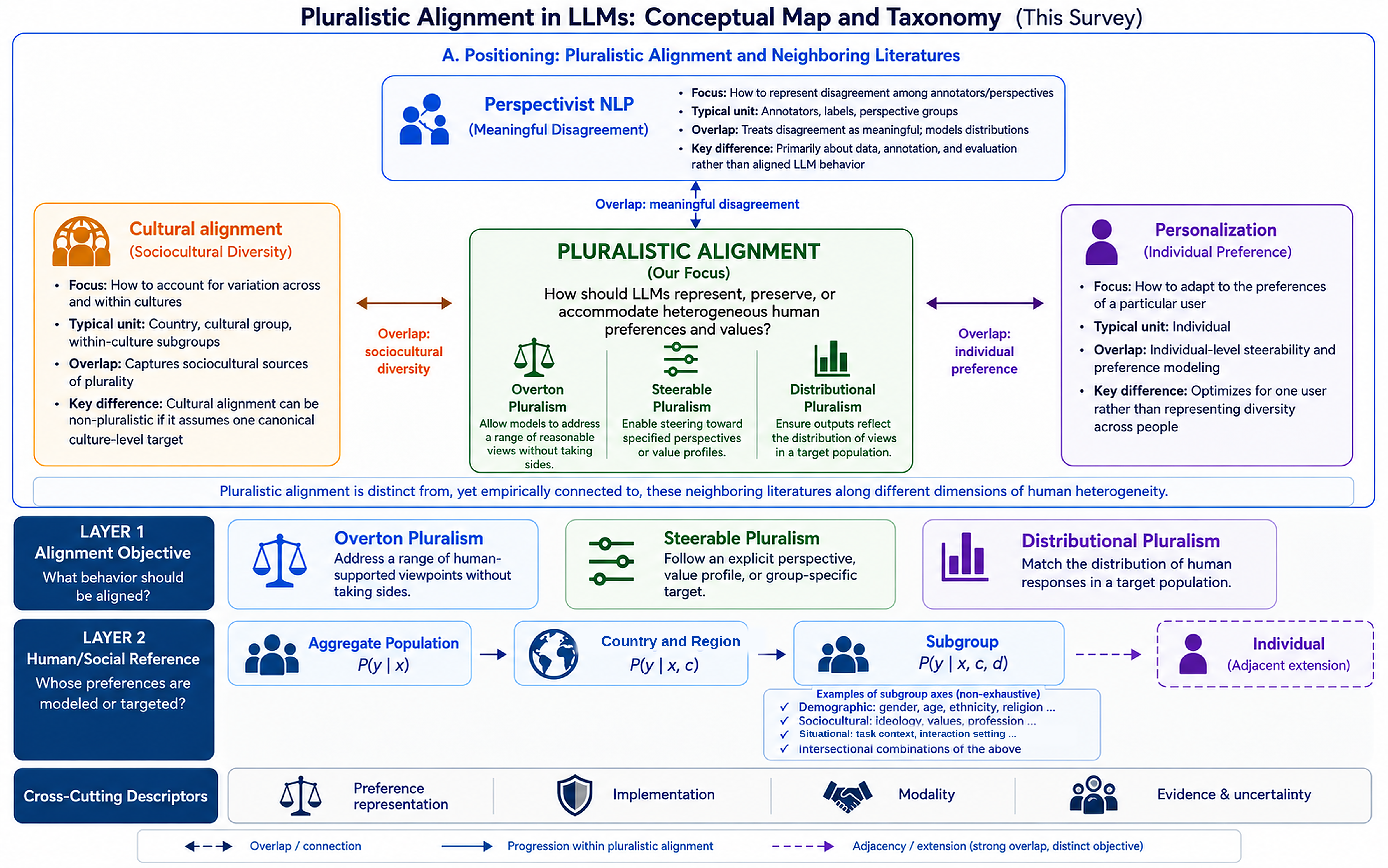}
\caption{\textbf{Conceptual map and two-layer taxonomy of pluralistic alignment.} Pluralistic alignment is positioned between neighboring literatures on perspectivist NLP, diversity-aware cultural alignment, and personalization. The socially grounded core taxonomy separates the pluralistic objective (\Opl, \Spl, or \Dpl) from the human reference (aggregate population, country and region, or subgroup). Individual personalization is an adjacent extension rather than a fourth human-reference category.}
\label{fig:framework}
\end{figure}

The framework proceeds from these two questions. We first define the three pluralistic objectives, then specify the human references relative to which they can be evaluated, clarify their relationship to neighboring research areas, and finally operationalize the framework as an objective-by-human-reference matrix. These two layers are intentionally narrower than the full system design space. Demographic attributes, feedback histories, value profiles, and latent representations determine how preference information is encoded, while prompting, parameter learning, and output aggregation determine how that information enters model behavior. Figure~\ref{fig:framework} summarizes these distinctions. Separating these dimensions allows methods that pursue the same target through different mechanisms to be compared without, for example, treating a synthetic persona as equivalent to an empirically measured population.

\subsection{Objective: What Should the Model Do?}

Layer~1 distinguishes three ways in which a model can respond to human disagreement. \textbf{\Opl} seeks to represent a range of human-supported perspectives on an issue. Its reference target is a set of viewpoints whose relevance or defensibility has been elicited or validated, rather than necessarily a frequency distribution. \textbf{\Spl} asks whether a model responds faithfully when conditioned on an explicit perspective, group, role, or value. The central criterion is fidelity to the specified condition, not simply whether the output changes. \textbf{\Dpl} instead asks whether the model reproduces the relative frequencies with which responses occur in a specified human reference:
\begin{equation}
P_M(y\mid x,g)\approx P_H(y\mid x,g),
\label{eq:distribution}
\end{equation}
where $x$ denotes a question, $y$ an answer, and $g$ the declared reference context. Because $P_H$ is estimated from human observations, the target distribution is itself generally uncertain rather than known exactly.

\begin{figure}[!htbp]
\centering
\includegraphics[width=\linewidth]{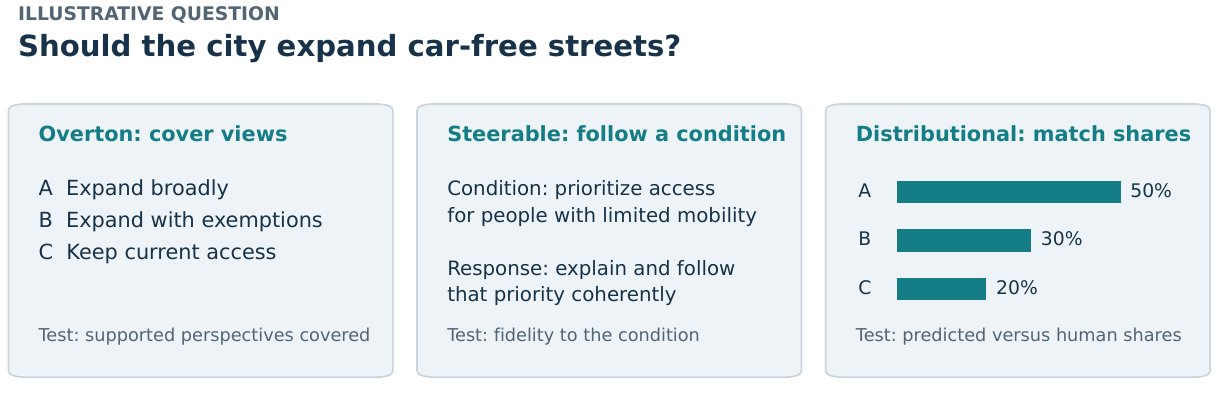}
\caption{One question, three evaluation targets. Evaluation for \Opl{} asks which defensible positions are covered; evaluation for \Spl{} asks whether an explicit condition is followed; evaluation for \Dpl{} asks whether response frequencies match the reference population. Positions and percentages are hypothetical. Neither the number of generated views nor their lexical diversity establishes all three properties.}
\label{fig:objectives}
\end{figure}

These objectives are complementary, and a single system may pursue more than one, as illustrated by Modular Pluralism and VITAL \citep{feng2024modular,shetty2025vital}. They nevertheless make distinct empirical claims and therefore require distinct forms of evidence. Generating many different answers does not guarantee coverage of human-supported positions, because high predictive entropy may reflect genuine disagreement or unsupported uncertainty. Similarly, responsiveness to a conditioning variable does not establish that the resulting behavior faithfully represents the intended perspective. Conversely, a concentrated distribution may be entirely appropriate when the reference population strongly agrees. Figure~\ref{fig:objectives} makes these distinctions concrete: perspective coverage, conditional fidelity, and frequency matching are related properties, but none serves as a substitute for the others.

\subsection{Human Reference: Whose Preferences Count?}

Once the desired behavior is specified, its empirical meaning depends on the people whose judgments define success. Layer~2 therefore identifies the human reference relative to which a Layer~1 objective is evaluated. We distinguish three operational classes. \textbf{Aggregate population} denotes a defined population evaluated without an explicit geographic or demographic partition; it does not necessarily refer to the global population. A reference population may itself be geographically bounded, such as a representative sample from a single country. We therefore use the \textbf{Country and Region} class only when geography is an explicit conditioning or comparison variable. \textbf{Subgroup} denotes an evaluated partition defined by one or more social or demographic attributes. A national average and an age-conditioned estimate derived from the same survey are thus different empirical targets even when they share the same sampling frame.

These categories organize empirical claims rather than rank identities by importance or specificity. Countries contain heterogeneous populations, while linguistic, religious, cultural, and other communities may span national boundaries. Granularity also introduces a statistical trade-off within the subgroup category. An age-by-gender estimate is not interchangeable with a gender marginal, and an age-by-gender-by-religion estimate may be supported by substantially fewer respondents. Finer reference groups are therefore informative only when the available data can support the corresponding estimand.

\begin{figure}[!htbp]
\centering
\includegraphics[width=\linewidth]{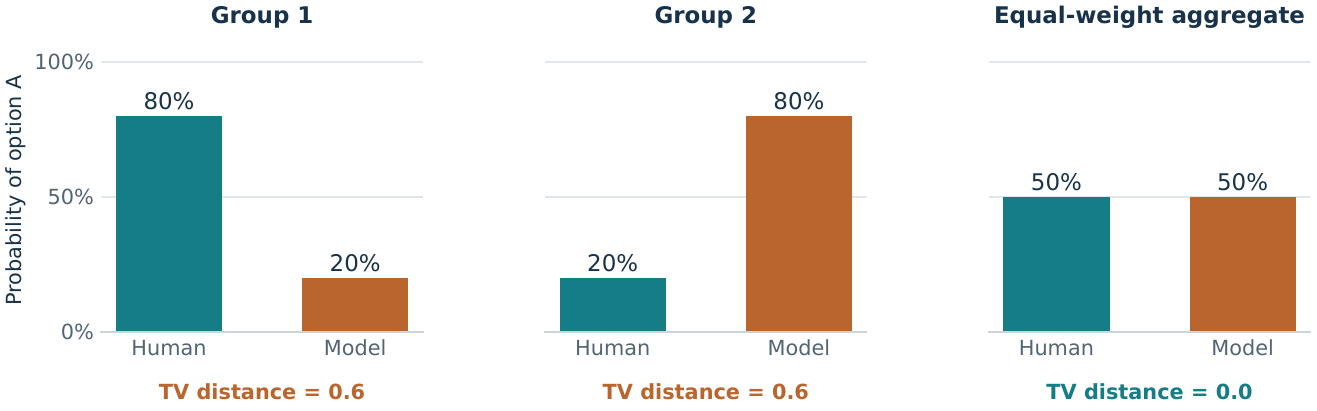}
\caption{Perfect aggregate agreement can hide large subgroup errors. In this hypothetical binary question, the two groups have equal weight. Humans prefer option A with probabilities 0.8 and 0.2; the model reverses them. Both aggregate distributions are $(0.5,0.5)$, giving total variation (TV) distance zero, while each subgroup has TV distance 0.6. The example establishes a mathematical possibility, not an empirical result from the corpus.}
\label{fig:aggregation}
\end{figure}

Aggregation mathematically connects different reference levels, but it does not make them interchangeable. When $g$ indexes a disjoint and exhaustive partition under a common sampling frame,
\begin{equation}
P_H(y\mid x)=\sum_g w_g(x)P_H(y\mid x,g),\qquad
w_g(x)=P_H(g\mid x).
\label{eq:mixture}
\end{equation}
The weights may be question-independent when the population definition and sampling conditions justify that assumption. Overlapping groups, however, cannot be combined by naively applying the same mixture formula. More importantly, agreement with the aggregate distribution does not identify agreement with its components. Overprediction for one subgroup may cancel underprediction for another, producing an accurate aggregate despite substantial subgroup-level errors. Figure~\ref{fig:aggregation} illustrates this possibility with a model that perfectly matches the aggregate distribution while reversing both subgroup distributions.

\paragraph{The estimand defines the evidence.}
The preceding example shows why the reference layer is more than descriptive metadata. A country-level claim concerns $P_H(y\mid x,c)$ for a specified question and reference frame, whereas a subgroup-level claim concerns $P_H(y\mid x,c,d)$ for grouping attributes $d$. A model can fit the former while averaging away substantial errors in the latter. Conversely, a model may respond strongly to a subgroup label while exaggerating differences that are not supported by observed respondents. The human-reference layer therefore determines both \emph{who} is being modeled and \emph{which empirical quantity} must be estimated to substantiate the corresponding alignment claim.

\subsection{Boundaries with Neighboring Literatures}

Separating objective from human reference also clarifies how pluralistic alignment relates to adjacent research areas. These literatures frequently share datasets, representations, and modeling techniques, but the claims they seek to establish are not identical.

\textbf{Perspectivist NLP treats human disagreement as meaningful information.}
Perspectivist approaches reject the assumption that every subjective instance admits a single gold label and instead preserve, model, or evaluate multiple human judgments \citep{davani2022disagreements,homan2022annotator}. This perspective ranges from retaining disaggregated annotations and annotator information to explicitly learning from disagreement during training, prediction, or evaluation. Pluralistic alignment shares this foundation but asks a different downstream question: how should a language model represent, expose, steer among, or reproduce heterogeneous human preferences in its behavior? The two areas are therefore complementary rather than equivalent. Perspectivist data and evaluation practices can provide the empirical foundation needed to study pluralistic behavior, while pluralistic alignment specifies what behavior a model should exhibit relative to those heterogeneous judgments. In particular, Distributional Pluralism additionally requires fidelity to the prevalence of responses in a defined human reference, rather than disagreement modeling alone.

\textbf{Cultural alignment provides context but does not guarantee pluralism.}
A model may reproduce what is presented as a country's dominant norm while overlooking substantial disagreement within that country. Diversity-aware cultural alignment makes the scope of cultural representation more explicit \citep{wu2025cultural}, but country, language, values, and practices should not be treated as interchangeable proxies for culture \citep{dev2026culture}. Cultural descriptors can help specify a reference context, but human observations remain necessary to establish what members of that reference population actually prefer.

\textbf{Personalization overlaps with pluralistic alignment without being equivalent to it.}
Personalization primarily asks whether a system's response is appropriate for a particular individual, whereas pluralistic alignment asks how multiple human preferences are represented, made accessible, or preserved in model behavior. The two areas may rely on similar profiles, prompts, feedback signals, or reward models, but success under one objective does not imply success under the other. Individual preference accuracy does not establish population-level coverage, just as population-level calibration does not guarantee an appropriate response for any particular user. We therefore treat individual-level modeling as an adjacent and interacting research direction rather than as an additional level in the human-reference hierarchy \citep{hwang2023useropinions,jiang2025individualistic,sorensen2025valueprofiles,poddar2024personalized}.

\subsection{The Objective-by-Human-Reference Matrix}

The two-layer taxonomy becomes operational through the matrix in Table~\ref{tab:joint-taxonomy}, which crosses the three pluralistic objectives with the three human-reference classes. Assignment to the matrix is intentionally conservative. A work must specify both a pluralistic objective and an empirically grounded human reference in its method or evaluation. A paper may occupy multiple cells when it separately evaluates multiple objectives or reference targets. By contrast, demographic attributes that appear only as components of an intersectional condition do not generate additional assignments.

Work centered on communities, roles, personas, values, or inferred latent groups remains part of the broader survey but enters the matrix only when its declared human target satisfies the corresponding reference criteria. This distinction prevents an unspecified or synthetically constructed perspective from being treated as an empirical population estimate. The appendices therefore report the complete coding decisions, inclusion criteria, and sensitivity analyses to make these assignments auditable.

\begin{table}[!htbp]
\centering\small
\caption{The objective-by-human-reference matrix for literature publicly available through October 14, 2026. Each cell gives its assignment count and direct citations. The 38 assignments come from 29 unique papers; rows and columns are not independent samples. Empty cells are bounded by the review protocol.}\label{tab:joint-taxonomy}
\renewcommand{\arraystretch}{1.18}
\begin{tabularx}{\linewidth}{@{}L{0.18\linewidth}XXX@{}}
\toprule
Objective & Aggregate population & Country and Region & Subgroup \\
\midrule
\Opl & \textbf{4 assignments}\par\smallskip\cite{fu2026opgrpo,mundada2026pluraleval,pooledayan2026overton,yang2026coverage} & \textit{No eligible work identified} & \textit{No eligible work identified} \\[7pt]
\Spl & \textit{No eligible work identified} & \textbf{7 assignments}\par\smallskip\cite{durmus2023globalopinions,yuan2024culturalpalette,xu2025culturespa,agarwal2026plural,borah2026pact,oh2026scpo,kotalawala2026plurva} & \textbf{9 assignments}\par\smallskip\cite{santurkar2023whose,zhao2024worldvaluesbench,chen2025spica,meister2025benchmarking,ali2026operationalizing,tan2026subgroup,zhu2026dvmap,feng2024modular,shetty2025vital} \\[7pt]
\Dpl & \textbf{3 assignments}\par\smallskip\cite{halpern2025calibrated,santurkar2023whose,russo2026moralgap} & \textbf{10 assignments}\par\smallskip\cite{durmus2023globalopinions,yuan2024culturalpalette,cao2025survey,luo2025sparse,borah2026pact,wu2026demographic,chae2026cultural,feng2024modular,shetty2025vital,yao2025gdpo} & \textbf{5 assignments}\par\smallskip\cite{santurkar2023whose,zhao2024worldvaluesbench,meister2025benchmarking,suh2025subpop,kambhatla2026distributional} \\[7pt]
\bottomrule
\end{tabularx}
\end{table}

The resulting matrix contains 29 unique papers and 38 paper-to-cell assignments. The remaining 41 works in the 70-work direct/adjacent corpus are documented outside the matrix in Appendix~\ref{app:outside}. Their exclusion reflects the matrix's explicit requirements for both objective and human reference, rather than a judgment about their importance to the broader literature. Accordingly, these counts characterize the works identified and coded under our protocol, not the prevalence of the corresponding objectives across the field as a whole.

Having specified \emph{what} is aligned and \emph{whose} preferences define success, we next turn to a separate design question: how a system represents the relevant human variation in the first place.

\section{Representing Human Diversity}\label{sec:representation}

The two-layer taxonomy specifies the alignment target, but it does not determine how the relevant people or preferences are represented in data or models. That choice shapes the evidence available to a system and, in turn, the claims its behavior can support. We organize representation choices along a progression from explicit social descriptors that identify constituencies, to observed judgments that record what people choose, to value profiles and latent models that infer structure when direct observations are sparse. We then consider how much resolution these representations can sustain before limited evidence is converted into artificial differences. These sources of information can be combined, but none should be conflated with the human target it is intended to approximate.

\subsection{Explicit Descriptors: Attributes, Communities, and Roles}

The most direct way to identify a reference group is through explicit descriptors. Countries and demographic attributes, for example, define constituencies for which human observations may be available. GlobalOpinionQA reports national opinion distributions, while OpinionQA and SubPOP support demographic or subpopulation analysis \citep{durmus2023globalopinions,santurkar2023whose,suh2025subpop}. Such descriptors make the unit of comparison transparent, but they do not explain why any particular individual holds a belief. Within-group variation can be substantial, and the meaning of a demographic label can shift across surveys and social settings. Explicit attributes are therefore most defensible as tools for defining and auditing a reference population, not as assumptions about a canonical group viewpoint.

Communities and roles extend this idea beyond demographic categories, but they encode different forms of structure. ComPO conditions on online communities, whereas healthcare pluralism distinguishes role-based viewpoints \citep{kumar2025compo,zhong2025healthcare}. A community may reflect shared practices, norms, or self-selection, while a role may encode situational responsibilities rather than a stable personal preference. Treating either as an ordinary demographic subgroup can therefore obscure what the condition actually represents. Evaluations should make this semantics explicit before interpreting conditioned behavior as evidence of fidelity to a human group.

Personas combine multiple descriptors into a more expressive profile. PERSONA uses demographic and idiosyncratic attributes, whereas MPTA constructs richer profiles from survey records \citep{castricato2025persona,tehenan2025mpta}. Such profiles can be useful for testing controllability, but their evidential value depends on provenance. A plausible synthetic persona is not equivalent to a sampled respondent, and generating many distinct personas does not estimate how prevalent those personas are in a population. The ability to elicit differentiated model behavior and the ability to represent empirical human diversity must therefore be evaluated separately.

\subsection{Observed Judgments: Direct Evidence of Preference}

Explicit descriptors indicate \emph{who} is being referenced, but observed judgments provide stronger evidence about \emph{what those people actually prefer}. Response histories, demonstrations, comparisons, and ratings record choices under specified conditions rather than relying solely on a model's prior associations with an identity label. PRISM links participant profiles to interaction-level feedback, and SPICA uses group-aware preference examples for in-context alignment \citep{kirk2024prism,chen2025spica}. Because these observations can vary within a nominal group, they also allow systems to preserve similarity when the data show little difference rather than forcing every group to appear distinct.

Observed feedback is only as informative as the elicitation process that produced it. A rating is conditional on the question and the candidate responses; engagement is not necessarily endorsement; and respondents cannot select viewpoints that never enter the candidate set. Community Alignment illustrates why candidate support matters for learning from heterogeneous preferences \citep{zhang2026community}. A dataset can therefore be diverse in its participants while remaining narrow in the opinions those participants are allowed to express. This distinction is especially important when observed choices are treated as a ground-truth preference distribution.

Repeated observations add another dimension by separating disagreement between people from change within a person. User histories and sequential resources can help distinguish persistent preferences from contextual or temporal drift \citep{sorensen2025valueprofiles,xu2026rushes}. Recent work on heterogeneous preference learning likewise models individuated utility functions instead of collapsing disagreement into a single reward \citep{mohan2026heterogeneous}. These targets are naturally individual-level, but their aggregate behavior can still be evaluated against population evidence when the sampling frame supports that interpretation.

\subsection{Inferred Representations: Values and Latent Structure}

When explicit descriptors or direct histories are unavailable, incomplete, or too sparse, systems may infer a lower-dimensional representation of preference. Value profiles summarize judgments in an interpretable form, while latent preference models infer components directly from heterogeneous feedback \citep{sorensen2025valueprofiles,siththaranjan2024dpl,chen2024pal}. These representations can capture structure that observable demographics miss, but they create an identification problem: a component that predicts comparisons well need not correspond to a stable value, community, or social group.

Predictive fit is therefore necessary but not sufficient for interpreting an inferred component as a human constituency. Useful checks include stability under resampling, generalization to new questions, and incremental predictive value beyond simple descriptors or a small number of prior answers. When value labels are attached to latent components, those interpretations should be validated with people rather than inferred from numerical separation alone. The same caution applies to natural-language profiles, mixtures of reward functions, and learned embeddings. A compact representation may be computationally useful without having a unique social interpretation.

\subsection{Choosing Resolution Without Inventing Differences}

Whether a representation is explicit, observed, or inferred, its resolution creates the same trade-off. Finer distinctions can reveal meaningful heterogeneity, but they also require stronger evidence. Combining attributes can expose differences hidden by marginal analyses, and WorldValuesBench and demographic-value work provide routes toward richer subgroup targets \citep{zhao2024worldvaluesbench,zhu2026dvmap,tan2026subgroup}. At the same time, each additional partition reduces the number of observations available per cell. Greater resolution is therefore useful only when the resulting distinctions are empirically supported.

A defensible analysis should justify finer partitions through both signal and uncertainty. Reporting cell sizes, comparing coarse and fine estimates on matched questions, and testing unseen attribute combinations can establish whether additional resolution is warranted. Hierarchical estimation or partial pooling can stabilize sparse groups, provided that the pooling assumptions are made explicit. Equally informative are questions on which groups respond similarly: a model that invents divergence on such items amplifies stereotypes rather than representing diversity. Faithful pluralism must preserve observed agreement as well as disagreement.

Crucially, representation choices do not define a form of pluralism by themselves. They determine what evidence is available for pursuing a particular objective with respect to a specified human reference. Once that evidence is represented, the next question is operational: \emph{where does it enter the model or system?}

\section{Implementing Pluralistic Behavior}\label{sec:methods}

The previous section considered how human heterogeneity is represented. We now ask how those representations shape model behavior. Implementation follows from the alignment target rather than defining it: the same conditioning, training, or composition mechanism can support different pluralistic objectives depending on the behavior it is intended to produce and the human reference against which that behavior is evaluated. We therefore organize methods by where preference information enters the system. Inference-time conditioning supplies labels, profiles, demonstrations, or retrieved judgments while keeping model parameters fixed. Learning-based methods internalize heterogeneous supervision in model parameters, reward functions, or preference models. Composition and calibration then determine how conditioned or learned components are combined into a final response or distribution. Finally, we consider multimodal systems, where the intervention logic remains similar but the evidence used to infer context becomes more complex. Table~\ref{tab:methods} summarizes the first three intervention points and their corresponding evaluation requirements.

\begin{table}[!htbp]
\centering\small
\caption{Three intervention locations with distinct evidential requirements. A method may use more than one.}\label{tab:methods}
\begin{tabularx}{\linewidth}{@{}L{0.18\linewidth}X X@{}}
\toprule
Intervention & Main design choice & Main evaluation question \\
\midrule
Condition the input & Labels, profiles, demonstrations, or retrieved feedback & Does the supplied condition improve fidelity to its human target? \\
Learn parameters & Distributional supervision, conditional rewards, or preference optimization & Does learned heterogeneity generalize beyond observed questions and groups? \\
Compose outputs & Select, combine, reweight, or calibrate components and candidates & Is relevant support present, and does the composition rule match the stated objective? \\
\bottomrule
\end{tabularx}
\end{table}

\subsection{Inference-Time Conditioning}

Inference-time conditioning is the lightest intervention because it changes the context presented to a fixed model while leaving its parameters untouched. The conditioning signal can range from coarse descriptors to direct observations. Country or demographic labels and culture-specific instructions provide explicit descriptors; value profiles and roles add richer semantic constraints; demonstrations and retrieved feedback ground the condition in observed judgments. This spectrum appears in group-aware in-context alignment such as SPICA \citep{chen2025spica}, instruction-driven approaches such as reasoning-based steering \citep{zhang2025cot} and culture-specific prompting \citep{xu2025culturespa}, and Demographic Pluralism \citep{wu-etal-2026-demographic}, which generates demographic-conditioned perspectives and aggregates their predictions. The last example illustrates that inference-time conditioning can support \Dpl{} as well as \Spl.

Conditioning is attractive because the target can change without retraining, but that flexibility also makes responsiveness easy to confuse with fidelity. A demographic phrase may trigger an exaggerated association, a lexical cue may dominate the intended condition, or steering may alter unrelated capabilities. Evaluation should therefore compare conditioned and unconditioned outputs against the \emph{same empirical target}, test robustness to paraphrases, and include questions on which reference groups exhibit little observed difference. These controls help distinguish faithful conditioning from stereotype amplification. A label alone is weak evidence for the distribution of opinions within a group. When preference information must persist across prompts or be learned from repeated supervision, training provides a more durable intervention.

\subsection{Learning Policies and Preference Models}

Training moves the preference signal from the input context into learned parameters, reward functions, or preference models. Methods differ primarily in the structure they seek to preserve. Survey-distribution specialization \citep{cao2025survey} and SubPOP \citep{suh2025subpop} optimize response distributions rather than only a majority answer. Group Preference Optimization \citep{zhao2024gpo} learns from group-specific feedback, whereas Distributional Preference Learning \citep{siththaranjan2024dpl}, PAL \citep{chen2024pal}, and GDPO \citep{yao2025gdpo} represent heterogeneity through preference or reward models. Other methods bind social conditions to richer targets: DVMap \citep{zhu2026dvmap} links demographic constraints to value-based alignment, and PlurVA-LLM \citep{kotalawala2026plurva} conditions on country across Chinese, Indonesian, and Sri Lankan settings. Multi-objective optimization provides a complementary form of steerability by exposing explicit trade-off weights rather than social-group conditions \citep{tsoi2026dial}.

Despite these differences, training-based approaches share a central identification problem: \emph{which aspects of human heterogeneity survive optimization}? Conditional rewards may preserve group differences, mixtures may preserve multiple preference functions, and distributional losses may preserve response frequencies. None can recover options that were never elicited, correct unrepresentative feedback, or repair a mislabeled reference group. Training therefore inherits both the strengths and the blind spots of the evidence used to define its objective.

Evaluation should consequently separate fit to observed strata from generalization beyond them. Useful splits hold out questions, groups, or attribute combinations and control overlap between related survey sources. A pooled average can conceal the group-specific failures that heterogeneous training is intended to address \citep{kambhatla2026distributional}. Latent models require evidence of component stability in addition to predictive accuracy, while explicit-group models require within-group calibration and checks against stereotype amplification. Even successful learning may leave several preference-sensitive components or candidate outputs, so producing a final prediction still requires an explicit composition rule.

\subsection{Composition, Aggregation, and Calibration}

Once a system contains preference-sensitive components, whether elicited at inference time or learned in parameters, it still needs a rule for turning them into a final prediction or response. Composition is this downstream decision point. Modular Pluralism makes the architecture explicit through specialized components \citep{feng2024modular}; Cultural Palette and role-based healthcare systems provide related examples \citep{yuan2024culturalpalette,zhong2025healthcare}. Modularity improves inspectability but moves the coverage problem upstream: downstream weighting cannot recover a constituency or perspective that is absent from the component set.

For distribution prediction, a common composition is
\begin{equation}
P_M(y\mid x)=\sum_{k=1}^{K}\alpha_k(x)P_k(y\mid x),\qquad
\alpha_k(x)\geq0,\quad\sum_k\alpha_k(x)=1.
\label{eq:model-mixture}
\end{equation}
The mixture is mathematically straightforward, but its social interpretation depends on the provenance of both the components and their weights. Population weighting estimates prevalence only when the components form a justified population partition and the weights correspond to that partition. Equal weighting is instead sensitive to how components are defined: splitting one perspective into two can increase its influence without any change in human opinion. Minority protection, population-frequency recovery, and worst-group welfare may therefore justify different weighting rules. Calibration can redistribute probability among represented answers, but it cannot create support for an omitted one.

This distinction also clarifies the relationship between pluralistic representation and consensus- or welfare-oriented alignment. Broad agreement, a collective constitution, and improved worst-group utility can each be valuable objectives \citep{bakker2022agreement,huang2024collective,chakraborty2024maxmin}, but none should be assumed to preserve the full distribution of disagreement. When these goals are combined, evaluation should report both the intended welfare benefit and any change in perspective coverage or frequency fidelity. The intervention structure itself is largely modality-agnostic; multimodal systems mainly change the evidence from which conditions and preferences are inferred.

\subsection{Modality Changes the Evidence, Not the Objective}

The intervention points above are easiest to see in text-only systems, but the same taxonomy extends to multimodal settings. Visual or audio inputs complicate the evidence used to infer context. Visual information may legitimately change which perspectives are relevant, but it may also encourage a model to infer identity or culture from superficial cues. Perspectivist and culturally situated vision-language studies make these risks visible \citep{viridiano2022perspective,karamolegkou2024vlm,yadav2025beyond}. They provide a foundation for multimodal pluralism, but cultural sensitivity in a vision-language task is not automatically evidence of \Opl, \Spl, or \Dpl.

Multimodal evaluation should therefore distinguish the people depicted in an input from the people whose judgments define the target, as well as explicit user conditions from context inferred by the model. Paired versions of the same scenario can test whether behavior tracks changes in human judgments or merely superficial modality cues. We therefore treat modality as a cross-cutting property rather than a fourth pluralistic objective. Across modalities and intervention points, the evaluation question is the same: \emph{what human evidence is sufficient to show that the claimed pluralistic behavior has been achieved?}

\section{Human Evidence and Evaluation}\label{sec:evaluation}

The preceding sections separate a pluralistic system into its target, its representation of human preferences, and the intervention used to shape behavior. Evaluation must reconnect each of these choices to human evidence. Following the objective layer, we ask whether \Opl{} covers valid perspective sets, whether \Spl{} moves behavior toward the intended condition rather than merely away from a default, and whether \Dpl{} matches empirical response frequencies while preserving credible support. We then consider requirements that cut across all three objectives, including disaggregation, generalization across groups and conditions, and uncertainty in the human reference. Table~\ref{tab:resource-guide} links these targets to representative resources, while Appendix~\ref{app:resources} provides a broader inventory. Throughout, we distinguish \emph{metadata that permit disaggregation} from \emph{observations sufficient to estimate a disaggregated target}: demographic fields alone do not guarantee the sample sizes required for reliable intersectional evaluation.

\begin{table}[!htbp]
\centering\small
\caption{Select resources according to the target being evaluated. Examples are representative rather than exhaustive; supporting a target does not imply that every paper using the resource evaluates it.}\label{tab:resource-guide}
\begin{tabularx}{\linewidth}{@{}L{0.19\linewidth}L{0.39\linewidth}X@{}}
\toprule
Human evidence needed & Representative resources & What to check before use \\
\midrule
Perspective sets & OvertonBench \citep{pooledayan2026overton}; WildSCOPE/PLURALEVAL \citep{mundada2026pluraleval} & Who supplied and validated the views; semantic matching rules \\
National frequencies & GlobalOpinionQA \citep{durmus2023globalopinions}; survey-distribution data \citep{cao2025survey} & Sampling frame, question wording, time, weights \\
Subgroup responses & OpinionQA \citep{santurkar2023whose}; SubPOP \citep{suh2025subpop}; WorldValuesBench \citep{zhao2024worldvaluesbench} & Cell sizes, available intersections, shared-source overlap \\
Interaction preferences & PRISM \citep{kirk2024prism}; SPICA data \citep{chen2025spica} & Rater identity, candidate support, repeated observations \\
Profiles and values & PERSONA \citep{castricato2025persona}; MPTA \citep{tehenan2025mpta}; Value Profiles \citep{sorensen2025valueprofiles} & Synthetic versus observed evidence; profile validity \\
\bottomrule
\end{tabularx}
\end{table}

\subsection{\Opl: Coverage Must Be Paired with Validity}

For \Opl, the central question is whether a model covers viewpoints that people actually support or regard as defensible. Let $\mathcal O_H(x)$ denote a human-supported reference set and $\mathcal O_M(x)$ the perspectives generated by the model. A simple conceptual decomposition is
\begin{equation}
\mathrm{Coverage}=\frac{|\mathcal O_M\cap\mathcal O_H|}{|\mathcal O_H|},\qquad
\mathrm{Validity}=\frac{|\mathcal O_M\cap\mathcal O_H|}{|\mathcal O_M|}.
\label{eq:coverage}
\end{equation}
These expressions assume a defined matching procedure and nonempty denominators. In open-ended text, the intersection is necessarily semantic rather than based on string equality. OvertonBench and PLURALEVAL provide concrete human-grounded approaches to constructing and evaluating such viewpoint sets \citep{pooledayan2026overton,mundada2026pluraleval}.

Coverage alone is insufficient because a system can appear more pluralistic simply by generating more views, including unsupported ones. Validity guards against this failure, while deduplication prevents paraphrases of the same position from being counted as distinct perspectives. At the same time, recall of low-frequency but human-supported positions can reveal omissions that an aggregate score would otherwise hide. The reference set is also socially situated: who proposed the viewpoints, who judged them relevant or defensible, and how disagreement among those judges was handled all shape the target being measured. Automated judges should therefore be validated against human assessments, particularly for perspectives unfamiliar to the judge.

A group-conditioned version of \Opl{} would make this dependence explicit by allowing $\mathcal O_g(x)$ to vary across human references rather than assuming one universally accepted set of viewpoints. Evaluation for \Opl{} thus asks whether a generated \emph{set} is both broad and grounded. \Spl{} changes the object of evaluation from a set of perspectives to behavior under an explicit condition.

\subsection{\Spl: Movement Is Not Fidelity}

\Spl{} shifts the evaluation question from \emph{which perspectives are present} to \emph{whether conditioning moves behavior toward the intended target}. A useful test therefore compares the conditioned model not only with its unconditioned behavior, but also with independent human evidence for the condition. When the target is a distribution, one diagnostic is
\begin{equation}
\Delta_g=D(P_0,P_H^g)-D(P_{s=g},P_H^g),
\label{eq:steering}
\end{equation}
where $P_0$ is unconditioned behavior, $P_{s=g}$ is behavior under condition $g$, and $D$ is a specified distributional discrepancy. A positive value indicates movement toward the human target rather than merely movement away from the default. When the condition denotes a role or value rather than an empirical response distribution, fidelity must instead be defined against evidence appropriate to that condition.

A convincing steering result also requires specificity and robustness. Improvement should persist under reasonable paraphrases of the condition, avoid collateral changes to unrelated capabilities, and leave questions with negligible empirical group differences largely unchanged. OpinionQA and GlobalOpinionQA illustrate why a named population requires an observed response target rather than a stereotype about that population \citep{santurkar2023whose,durmus2023globalopinions}. Responsiveness shows that a model can be steered; only agreement with independently grounded evidence shows that it is steered \emph{faithfully}. A stronger claim, namely that the model reproduces the full pattern of human response frequencies, requires evaluation for \Dpl.

\subsection{\Dpl: Match Frequencies and Retain Support}

\Dpl{} requires a different form of evidence: the full allocation of probability mass must be compared with an empirical human distribution. For categorical distributions $P$ and $Q$, commonly used discrepancies include Jensen--Shannon (JS) divergence and total variation (TV) distance,
\begin{align}
\mathrm{JS}(P,Q)&=\tfrac12\mathrm{KL}(P\|M)+\tfrac12\mathrm{KL}(Q\|M),\quad M=\tfrac12(P+Q),\\
\mathrm{TV}(P,Q)&=\tfrac12\sum_y|P(y)-Q(y)|.
\end{align}
We use $\mathrm{JS}$ to denote the divergence; its square root is the Jensen--Shannon \emph{distance}. Reports should state which quantity and logarithm base they use. Unlike JS, $\mathrm{KL}(P\|Q)$ can become infinite when $P(y)>0$ but $Q(y)=0$, so the direction of KL is consequential.

A single distributional discrepancy, however, does not reveal every relevant failure. A model can achieve a small TV or JS value while assigning zero probability to a low-frequency but credible answer. Conversely, it can retain support for every answer while allocating the wrong frequencies. Distributional fidelity therefore benefits from pairing an overall discrepancy with diagnostics for missing support and minority-tail behavior, ideally disaggregated by the relevant human reference. Cross-country settings add a relational concern: even when each country's marginal distribution appears individually close to its target, a model may compress or exaggerate the differences between countries. Cultural Divergence Preservation makes this failure mode explicit by measuring cultural flattening and caricature alongside conventional fidelity metrics \citep{chae2026cultural}. Log loss and Brier scores provide additional probability-quality checks when suitable human observations are available, whereas correlation with survey percentages alone cannot establish that the frequencies themselves are matched.

The estimated model distribution also depends on how responses are elicited. First-token option probabilities, sampled answers, and self-reported percentages are distinct observables and should not be treated as interchangeable \citep{meister2025benchmarking}. Studies should report option tokenization and ordering, normalization, temperature, prompt format, and sampling count where relevant. The same caution applies to the human side of the comparison: empirical histograms have sampling error, so uncertainty in the reference distribution should be propagated rather than treating every observed percentage as exact.

\paragraph{Illustrative diagnostic.}
Consider an item with three plausible answers for which respondents in one subgroup choose $(0.7,0.2,0.1)$. A system that predicts $(0.7,0.3,0.0)$ moves only 0.1 probability mass and therefore has TV distance 0.1, yet it assigns zero support to an answer selected by one in ten respondents. That failure is qualitatively different from a similarly small frequency error that preserves support for all three responses. A careful evaluation should therefore report both aggregate discrepancy and behavior on low-frequency but credible responses, together with uncertainty intervals when those support estimates come from few observations. The example is hypothetical; its purpose is to show why a single divergence score does not identify which opinions a model overlooks.

\paragraph{Reference uncertainty.}
Human response proportions are estimates rather than fixed truths. Their uncertainty depends on effective sample size, survey weighting, question wording, response order, and the time and context of collection. A seemingly precise subgroup target may be dominated by sampling noise or by the way a small group was defined. Cross-country comparisons can additionally confound genuine preference differences with translation or survey-mode effects. Researchers should therefore document these conditions, avoid reporting cell-level distances without the associated denominators, and, where feasible, repeat conclusions under plausible alternative weighting or grouping schemes. This is especially important when reported performance differences are smaller than the uncertainty in the human reference itself. These diagnostics establish whether a particular objective is met for a specified target, but they do not yet show whether performance is reliable across multiple groups, conditions, or time periods.

\subsection{Evaluate Across Groups and Conditions}

The preceding diagnostics are objective-specific; the human-reference layer adds a set of cross-cutting requirements. Most importantly, aggregate and subgroup results should be interpreted together. Figure~\ref{fig:aggregation} shows why: micro-averaging emphasizes frequent observations, macro-averaging gives evaluated cells equal weight, and worst-group reporting exposes concentrated failures. None is universally correct. The averaging rule should follow the claimed estimand, and estimates for small groups should be accompanied by uncertainty. Survey weights and respondent clustering should also be respected when required by the underlying data.

Generalization must likewise be defined with respect to the reference structure. Holding out new questions tests a different capability from holding out groups, unseen attribute combinations, or later collection periods. Success on age and gender separately, for example, does not establish success on age-by-gender intersections. Shared survey sources can also create hidden overlap between nominally distinct benchmarks. These are properties of the evaluation design, and a higher pooled score does not resolve them.

Finally, behavioral fidelity should not be conflated with explanatory fidelity. Matching answer frequencies does not show that a model has recovered the reasons people give for those answers. A method that claims faithful rationales therefore needs evidence about rationales in addition to response distributions. Perspectivist evaluation is useful here because it can retain disagreement among human judges instead of replacing it with a single adjudicated label \citep{davani2022disagreements,homan2022annotator}. This distinction is especially important when the evaluation concerns contested concepts such as reasonableness or cultural appropriateness.

\paragraph{A minimum reporting standard.}
Taken together, these considerations suggest a compact reporting discipline. Every pluralistic-alignment result should state the claimed objective, the human reference and sampling frame, how judgments and candidate responses were collected, how model responses or probabilities were elicited, and which metric tests the claim. Subgroup conclusions should report cell sizes and uncertainty alongside aggregate scores. The goal is not to impose a universal metric, but to ensure that the reported evidence directly bears on the behavior being claimed. With these requirements in place, the objective-by-human-reference matrix becomes more than a catalogue of methods: it reveals which pluralistic claims are well supported and which remain under-evaluated.

\section{What the Objective-by-Human-Reference Matrix Shows}\label{sec:landscape}

Having separated the alignment objective, human reference, preference representation, implementation mechanism, and evaluation requirements, we can now examine where existing work provides empirical support across this design space. The objective-by-human-reference matrix serves as the primary lens, while the preceding sections help interpret what its occupied and empty cells establish. We first summarize three patterns in the coded corpus and then translate those patterns into research priorities.

The matrix contains 29 of the 70 direct or adjacent works identified under our inclusion protocol, yielding 38 paper-to-cell assignments because a paper may occupy multiple cells. Appendix~\ref{app:protocol} details the search and coding protocol, Appendix~\ref{app:matrix-manifest} provides the full objective-by-human-reference matrix coding, and Appendix~\ref{app:direct-manifest} documents the complete direct and adjacent corpus.

\subsection{Three Findings and Their Limits}

Three patterns stand out when the coded corpus is viewed through the objective-by-human-reference matrix.

\textbf{First, the identified literature is weighted toward \Spl{} and \Dpl.} Of the 38 assignments, 4 concern \Opl, 16 concern \Spl, and 18 concern \Dpl. These are assignment counts rather than counts of independent papers because a paper may address more than one objective. The same broad pattern appears in the peer-reviewed subset (2, 10, 12 assignments) and the high-confidence subset (3, 16, 18), reported in Appendix~\ref{app:sensitivity}. The imbalance describes the composition of our protocol-bounded corpus; it does not establish that one research direction is more mature, effective, or important than another.

\textbf{Second, subgroup-level evaluation is substantial but less prevalent under target-based coding.} The subgroup column contains 14 assignments, compared with 17 for country and region and 7 for aggregate population. This distinction matters because demographic attributes can appear in a method without defining its empirical target. We assign a subgroup cell only when subgroup-specific judgments or distributions are themselves evaluated; demographic cells used only for conditioning, aggregation, or sample balancing do not qualify. Even among retained subgroup assignments, the matrix does not indicate whether a paper evaluates a single demographic marginal, several intersections, or groups supported by adequate sample sizes. The substantive question is therefore not simply whether demographic variables appear, but whether subgroup-level estimands are explicit and statistically reliable.

\textbf{Third, the clearest empty cells involve socially conditioned \Opl.} Every eligible \Opl{} assignment in the matrix uses an aggregate-population reference; under our criteria, no country- or subgroup-conditioned \Opl{} paper qualifies. This bounded absence points to a concrete measurement problem: which perspectives are broadly supported across groups, which are supported primarily within particular groups, and can a model communicate both facts without erasing attribution? The empty cell for \Spl{} at the aggregate-population level is more sensitive to coding conventions because generic perspective steering does not necessarily specify an empirical population. These empty cells are therefore best interpreted as prompts for further measurement, not as proof that the corresponding problems have never been studied.

These patterns should be interpreted as properties of our protocol-bounded corpus rather than estimates of the field as a whole. The matrix intentionally excludes work that does not specify both an eligible pluralistic objective and an empirically grounded human reference. The appendices provide paper-level coding decisions and sensitivity analyses to make these judgments auditable.

\subsection{Priorities Derived from the Framework}

The descriptive patterns are most useful when translated into research-design questions. The priorities below follow the paper's progression: strengthen the human reference, compare representation choices, make composition rules explicit, extend evaluation to interaction and multimodality, and build benchmarks for sparsely studied cells in the objective-by-human-reference matrix. Across these directions, the central question is not simply how much outputs vary, but whose perspectives are represented, what evidence supports them, and how faithfully the system preserves that evidence.

\paragraph{Build human references that preserve disagreement.}
Population averages should be accompanied by respondent-level or appropriately disaggregated evidence whenever access, privacy, and licensing permit. Datasets for \Opl{} have an additional requirement: they should record who proposed and validated each viewpoint, including disagreement over what counts as relevant or defensible. Candidate collection should also seek perspectives that are absent from common model outputs. Otherwise, even a demographically diverse rater pool cannot reveal omissions from the candidate space.

\paragraph{Compare representation choices under shared conditions.}
The representation choices in Section~\ref{sec:representation}, including demographic descriptors, observed judgments, values, and latent factors, should be compared on matched questions and comparable training budgets. The empirical question is what each representation adds, for which people, and on which held-out examples, rather than whether a richer profile merely appears culturally plausible. Coverage audits should examine participants, task types, and candidate answers separately \citep{wu2025cultural,dev2026culture}.

\paragraph{Make aggregation choices explicit.}
The composition analysis in Section~\ref{sec:methods} implies that estimating population prevalence, giving each perspective equal visibility, improving minority access, and maximizing worst-group welfare are not interchangeable goals. A prevalence estimator may require justified survey weights, whereas a deliberation assistant may intentionally foreground neglected perspectives. Both choices can be defensible, but evaluation should state whether the human distribution is being estimated or deliberately transformed. Reporting that distinction is more informative than treating every welfare improvement as a generic gain in pluralism.

\paragraph{Extend evaluation to interaction and multimodality.}
Static response matching cannot establish whether users can discover alternative views, steer a system predictably, or correct mistaken assumptions about them. Interaction studies can test these properties directly \citep{pooledayan2026impact}. Multimodal evaluation introduces an additional distinction between the identity of people depicted in an input and the identity of those whose judgments define the target. Matched scenarios and repeated judgments can help determine whether behavior changes because the relevant human reference changed or merely because visual or linguistic cues changed.

\paragraph{A concrete benchmark direction: group-conditioned \Opl.}
One immediate opportunity is to combine the sampling discipline of distributional surveys with the viewpoint-set evaluation used for \Opl. A benchmark could present the same contested question to several well-sampled groups, elicit multiple candidate perspectives, and record which perspectives each group regards as relevant or defensible. Retaining both endorsements and disagreement about reasonableness would support group-specific coverage and validity measures, as well as comparisons of overlap among viewpoint sets. Crucially, a stereotypical position should not receive credit merely because it is associated with a demographic label; the perspective must be supported by respondents from the declared reference.

\paragraph{Separate descriptive fidelity from normative choice.}
A final priority cuts across all three objectives. Learning what people believe is different from deciding how a system should act on those beliefs. A model estimating public opinion aims to preserve an empirically justified distribution; a deliberative tool may intentionally emphasize underrepresented arguments; a fairness-oriented policy may prioritize the worst-served group. These systems need not share a single optimum. Making the intended transformation explicit would clarify their consequences and prevent descriptive alignment from being conflated with a particular normative aggregation rule.

\section{Conclusion}

Pluralistic alignment becomes more precise when two questions are stated explicitly: \emph{how should a model respond to human disagreement, and whose judgments should define success}? Our framework addresses them through two complementary dimensions. The alignment objective distinguishes \Opl, \Spl, and \Dpl, while the human reference identifies the aggregate population, country and region, or subgroup whose judgments ground evaluation. Together, these dimensions clarify both the behavior being claimed and the evidence needed to assess it. Preference representations determine what variation a system can access; implementation mechanisms determine how that variation shapes behavior; and evaluation tests fidelity to the intended human reference. The literature also reveals a concrete opportunity for group-conditioned \Opl, alongside persistent challenges in sampling, uncertainty, aggregation, and generalization. More broadly, pluralistic alignment should be grounded in identifiable human evidence and explicit about the trade-offs that arise when preferences conflict.

\clearpage
\begingroup\small
\setlength{\bibsep}{4pt plus 1pt}
\bibliographystyle{plainnat}
\bibliography{references}
\endgroup
\clearpage
\appendix
\section{Full Search and Coding Protocol}\label{app:protocol}
This appendix reproduces the finalized protocol used for literature coverage through October 14, 2026. The objective-by-human-reference matrix targets \emph{socially grounded pluralistic alignment}, rather than every pluralistic representation or aggregation mechanism.

\subsection{Search Families}
The corpus was assembled iteratively through keyword search, benchmark/method-name search, and backward/forward citation tracing. Search families combined language-model or LLM terms with:
\begin{enumerate}[leftmargin=1.5em]
  \item ``pluralistic alignment'', ``Overton Pluralism'', ``Steerable Pluralism'', and ``Distributional Pluralism'';
  \item ``distributional alignment'' + human preferences, ``heterogeneous preferences'' + alignment, and ``group preference'' + alignment;
  \item ``survey response distributions'' + language model, subpopulation public opinion + LLM, and demographic alignment + LLM;
  \item ``pluralistic alignment'' combined with DPO, federated learning, persona, and value terms;
  \item cultural preference alignment + LLM, individualistic values + LLM, and value profiles + LLM; and
  \item multimodal pluralism, multimodal cultural alignment, and vision-language cultural preference.
\end{enumerate}
Primary or authoritative records were checked when available in ACL Anthology, PMLR, OpenReview, AAAI proceedings, arXiv, and official project pages. Duplicate preprint, workshop, and venue versions were consolidated to the most authoritative version available by the coverage cutoff.

\subsection{Targeted Supplementary Search}
A targeted high-recall supplementary sweep on August 19, 2026 reran direct queries for method families and exact titles, including ``Steerable Pluralism: Few-Shot Comparative Regression'', ``MixDPO: Pluralistic Alignment'', ``Coherence Maximization Improves Pluralistic Alignment'', and ``PluralLLM'' + federated pluralistic alignment. Venue- and site-restricted searches were also run over AAAI, ACL Anthology, PMLR, ICML/NeurIPS publication records, and official laboratory publication pages, followed by exact-title verification of candidate records. The supplementary searches identified direct or adjacent studies such as Steerable Pluralism, MixDPO, Coherence Maximization, PluralLLM, FGD-Align, ALPHA, DisAlign, Demo-SafetyBench, LLMs' Pluralistic Compatibility, PICACO, and COUPLE. Rushes was retained as an adjacent study of persistent user-level preferences. The resulting records, including individual- and persona-defined work outside the objective-by-human-reference matrix, appear in Appendix~\ref{app:direct-manifest}.

A delta update on October 14, 2026 reran the same search families for work made public after August 19. Applying the unchanged inclusion and coding rules added five publicly available studies to the broader corpus. Three meet the objective-by-human-reference matrix criteria: Demographic Pluralism \citep{wu-etal-2026-demographic}, Cultural Divergence Preservation \citep{chae2026cultural}, and a PlurVA-LLM shared-task system on multilingual, country-conditioned value alignment \citep{kotalawala2026plurva}. Two additional studies are retained as adjacent boundary cases: steerable multi-objective trade-off analysis \citep{tsoi2026dial} and individuated heterogeneous-preference learning \citep{mohan2026heterogeneous}. This staged update extends temporal coverage; final matrix coding was subsequently re-audited under the operational objective and human-reference rules below.

\subsection{Screening Stages}
\begin{enumerate}[leftmargin=1.5em]
  \item \textbf{Title/abstract screen.} Retain records explicitly about heterogeneous human preferences, pluralistic behavior, or reusable human-disagreement resources.
  \item \textbf{Full-record relevance screen.} Classify each retained record as direct pluralistic work, supporting literature, or an adjacent objective/boundary case; direct and adjacent retained records together form the broader direct/\allowbreak{}adjacent corpus.
  \item \textbf{Objective-by-Human-Reference Matrix eligibility screen.} A paper enters the 3-by-3 matrix only when both (a) an objective corresponding to \Opl, \Spl, or \Dpl{} and (b) an explicit human-reference level are present in its design or evaluation.
\end{enumerate}
The iterative snowballing protocol does not yield a single meaningful initial database-hit denominator, so we do not report a pseudo-PRISMA retrieval count. Instead, we report exact retained-corpus counts, search families, eligibility rules, and supplementary searches used to increase recall.

\subsection{Human-Reference Coding Rules}
\begin{itemize}[leftmargin=1.5em]
  \item \textbf{Aggregate population:} a named or operational human reference population is modeled without an explicit social partition. Aggregate population is \emph{not} a residual bucket for value-, persona-, community-, latent-, or individual-defined work.
  \item \textbf{Country and Region:} geography is an explicit conditioning or comparison variable.
  \item \textbf{Subgroup:} one or more explicit social, demographic, or sociocultural attributes define the target below the aggregate-population or country-and-region level.
  \item Value vectors, abstract perspectives, communities, roles, personas, latent clusters, method-defined preference groups, and individuals remain outside the objective-by-human-reference matrix unless the paper separately evaluates an explicit human reference level.
  \item Human-reference categories are assigned from the empirical target used to support the alignment claim, not from intermediate representations or sampling controls. Demographic cells used only for conditioning, aggregation, or sample balancing do not constitute subgroup-level evaluation.
  \item A dataset's capacity to support subgroup analysis is not sufficient; the paper must directly state or evaluate that target.
  \item Intersectional targets are coded at the level of the full evaluated reference. For example, a continent-by-urbanicity-by-education target is coded as Subgroup rather than also receiving a Country and Region assignment, unless the paper separately evaluates a geography-only target. A paper may occupy multiple human-reference columns when it separately evaluates distinct targets, such as both an aggregate population and demographic subgroups.
  \item Empty cells mean that no direct work satisfying the strict cell definition was identified under the stated protocol and coverage window; they are not universal absence claims.
\end{itemize}

\subsection{Objective Coding Rules}
The matrix uses the survey's operational definitions rather than paper titles or method labels alone. This keeps objective coding tied to what is actually evaluated.
\begin{itemize}[leftmargin=1.5em]
  \item \textbf{\Opl:} the evaluated target is coverage of a human-supported viewpoint set, together with evidence that generated viewpoints remain within that supported set. Coverage or support breadth does not become \Dpl{} unless response frequencies are also modeled.
  \item \textbf{\Spl:} the model is conditioned on an explicit human perspective, group, country, role, or value target and is evaluated for fidelity to that condition. Merely including a demographic or geographic descriptor in a prompt or training record is insufficient.
  \item \textbf{\Dpl:} model probability mass or empirical response frequencies are compared with an observed human response distribution for the declared reference. Diversity, entropy, or coverage without frequency matching is not sufficient.
  \item When a paper's terminology differs from these operational definitions, the matrix follows the evaluated target and records the ambiguity in the coding note.
\end{itemize}

\subsection{Corpus and Sensitivity Sets}
The direct/\allowbreak{}adjacent corpus contains \textbf{70 unique papers}: \textbf{29 papers eligible for the objective-by-human-reference matrix} and \textbf{41 direct/\allowbreak{}adjacent papers outside the objective-by-human-reference matrix}. The matrix papers yield \textbf{38 paper-to-cell assignments}: 4 for \Opl, 16 for \Spl, and 18 for \Dpl; by human reference, 7 concern aggregate populations, 17 country and region, and 14 subgroups. The sensitivity sets are: all socially grounded (29 papers/38 assignments), empirical-only (29/38), peer-reviewed-only (18/24), and high-confidence-only (28/37).

The review is protocol-bounded and single-coded, not a formal PRISMA review with independent duplicate coders; no inter-rater agreement is claimed. Matrix records therefore label \emph{Coding Confidence} as either High or Borderline. High confidence requires an explicit objective corresponding to \Opl, \Spl, or \Dpl{} and an explicit Aggregate population, Country and Region, or Subgroup reference in the method or evaluation under the operational rules above. Borderline is used when the empirical target is eligible but its objective admits a plausible alternative reading. In the current coding, the coverage framework of \citet{yang2026coverage} is Borderline because the authors describe \Dpl, whereas its reported coverage and in-boundary metrics fit our operational definition of \Opl. PRISM and PERSONA remain outside the matrix because individualized or synthetic-persona evidence does not by itself establish an aggregate, country-level, or subgroup target for \Opl, \Spl, or \Dpl.

\section{Full Objective-by-Human-Reference Matrix Corpus Coding}\label{app:matrix-manifest}
The following two companion tables provide the complete paper-level coding used for the objective-by-human-reference matrix. Part I reports bibliographic and contribution information; Part II reports human-reference, matrix-assignment, and coding-confidence information. The \emph{Reference} column uses the paper's numbered citation in the bibliography, which also links the two views of each paper.

\scriptsize
\setlength{\tabcolsep}{3pt}
\renewcommand{\arraystretch}{1.08}
\begin{longtable}{@{}L{0.52in}L{0.40in}L{3.38in}L{1.02in}L{0.78in}@{}}
\caption{Objective-by-Human-Reference Matrix corpus coding, Part I: bibliographic and contribution information.}\label{tab:matrix-manifest-a}\\
\toprule
\textbf{Ref.} & \textbf{Year} & \textbf{Title} & \textbf{Contribution type} & \textbf{Venue status} \\
\midrule
\endfirsthead
\multicolumn{5}{l}{\tablename\ \thetable\ (continued)}\\
\toprule
\textbf{Ref.} & \textbf{Year} & \textbf{Title} & \textbf{Contribution type} & \textbf{Venue status} \\
\midrule
\endhead
\midrule
\multicolumn{5}{r}{Continued on next page}\\
\endfoot
\bottomrule
\endlastfoot
\cite{durmus2023globalopinions} & 2023 & Towards Measuring the Representation of Subjective Global Opinions in Language Models & Dataset/\allowbreak{}Benchmark & preprint \\
\cite{santurkar2023whose} & 2023 & Whose Opinions Do Language Models Reflect? & Dataset/\allowbreak{}Benchmark & peer-reviewed \\
\cite{yuan2024culturalpalette} & 2024 & Cultural Palette: Pluralising Culture Alignment via Multi-agent Palette & Method & preprint \\
\cite{zhao2024worldvaluesbench} & 2024 & WorldValuesBench: A Large-Scale Benchmark Dataset for Multi-Cultural Value Awareness of Language Models & Dataset/\allowbreak{}Benchmark & peer-reviewed \\
\cite{cao2025survey} & 2025 & Specializing Large Language Models to Simulate Survey Response Distributions for Global Populations & Method & peer-reviewed \\
\cite{chen2025spica} & 2025 & SPICA: Retrieving Scenarios for Pluralistic In-Context Alignment & Method & peer-reviewed \\
\cite{halpern2025calibrated} & 2025 & Pairwise Calibrated Rewards for Pluralistic Alignment & Method & preprint \\
\cite{luo2025sparse} & 2025 & Towards Low-Resource Alignment to Diverse Perspectives with Sparse Feedback & Method & peer-reviewed \\
\cite{meister2025benchmarking} & 2025 & Benchmarking Distributional Alignment of Large Language Models & Dataset/\allowbreak{}Benchmark & peer-reviewed \\
\cite{suh2025subpop} & 2025 & Language Model Fine-Tuning on Scaled Survey Data for Predicting Distributions of Public Opinions & Method & peer-reviewed \\
\cite{xu2025culturespa} & 2025 & Self-Pluralising Culture Alignment for Large Language Models & Method & peer-reviewed \\
\cite{agarwal2026plural} & 2026 & PLURAL: A Global Dataset for Value Alignment & Dataset/\allowbreak{}Benchmark & preprint \\
\cite{ali2026operationalizing} & 2026 & Operationalizing Pluralistic Values in Large Language Model Alignment Reveals Trade-offs in Safety, Inclusivity, and Model Behavior & Empirical Analysis & peer-reviewed \\
\cite{borah2026pact} & 2026 & Whose Norms? Disentangling Cultural and Personal Alignment in Large Language Models & Empirical Analysis & preprint \\
\cite{fu2026opgrpo} & 2026 & Overton Pluralistic Reinforcement Learning for Large Language Models & Method & preprint \\
\cite{kambhatla2026distributional} & 2026 & Improving the Distributional Alignment of LLM & Method & peer-reviewed \\
\cite{mundada2026pluraleval} & 2026 & Evaluating Language Model Pluralism through In-the-wild Crowd Discussions & Dataset/\allowbreak{}Benchmark & peer-reviewed \\
\cite{oh2026scpo} & 2026 & Steerable Cultural Preference Optimization of Reward Models & Method & preprint \\
\cite{pooledayan2026overton} & 2026 & Benchmarking Overton Pluralism in LLMs & Dataset/\allowbreak{}Benchmark & peer-reviewed \\
\cite{russo2026moralgap} & 2026 & The Pluralistic Moral Gap: Understanding Moral Judgment and Value Differences between Humans and Large Language Models & Empirical Analysis & peer-reviewed \\
\cite{tan2026subgroup} & 2026 & Can Persona-Prompted LLMs Emulate Subgroup Values? An Empirical Analysis of Generalisability and Fairness in Cultural Alignment & Empirical Analysis & peer-reviewed \\
\cite{yang2026coverage} & 2026 & Where Models Converge and Humans Diverge: A Coverage Framework for Distributional Pluralism in Open-Ended Generation & Evaluation Framework & preprint \\
\cite{zhu2026dvmap} & 2026 & DVMap: Fine-Grained Pluralistic Value Alignment via High-Consensus Demographic-Value Mapping & Method & peer-reviewed \\
\cite{feng2024modular} & 2024 & Modular Pluralism: Pluralistic Alignment via Multi-LLM Collaboration & Method & peer-reviewed \\
\cite{shetty2025vital} & 2025 & VITAL: A New Dataset for Benchmarking Pluralistic Alignment in Healthcare & Dataset/\allowbreak{}Benchmark & peer-reviewed \\
\cite{yao2025gdpo} & 2025 & No Preference Left Behind: Group Distributional Preference Optimization & Method & peer-reviewed \\
\cite{wu-etal-2026-demographic} & 2026 & Demographic Pluralism: Inference-Time Modeling of Pluralistic Human Preference Distributions & Method & preprint \\
\cite{chae2026cultural} & 2026 & Cultural Divergence Preservation: Diagnosing Flattening and Caricature in LLM-Simulated Survey Populations & Evaluation Framework & workshop \\
\cite{kotalawala2026plurva} & 2026 & PlurVA-LLM-2026 Shared Task Track-1: Pluralistic Value Alignment in LLMs via Multilingual Fine-Tuning and Threshold Calibration & Method/\allowbreak{}System & workshop \\
\end{longtable}
\normalsize

\scriptsize
\setlength{\tabcolsep}{3pt}
\renewcommand{\arraystretch}{1.08}
\begin{longtable}{@{}L{0.52in}L{0.86in}L{2.36in}L{0.68in}L{1.54in}@{}}
\caption{Objective-by-Human-Reference Matrix corpus coding, Part II: human-reference, assignment, and coding-confidence information.}\label{tab:matrix-manifest-b}\\
\toprule
\textbf{Ref.} & \textbf{Human reference} & \textbf{Matrix assignments} & \textbf{Confidence} & \textbf{Confidence note} \\
\midrule
\endfirsthead
\multicolumn{5}{l}{\tablename\ \thetable\ (continued)}\\
\toprule
\textbf{Ref.} & \textbf{Human reference} & \textbf{Matrix assignments} & \textbf{Confidence} & \textbf{Confidence note} \\
\midrule
\endhead
\midrule
\multicolumn{5}{r}{Continued on next page}\\
\endfoot
\bottomrule
\endlastfoot
\cite{durmus2023globalopinions} & Country and Region & \Spl{}: Country and Region; \Dpl{}: Country and Region & High & GlobalOpinionQA conditions on a target country and compares model answer distributions with country-specific survey distributions, supporting both country-level steering and distributional evaluation. \\
\cite{santurkar2023whose} & Aggregate population; Subgroup & \Spl{}: Subgroup; \Dpl{}: Aggregate population; \Dpl{}: Subgroup & High & OpinionQA evaluates the overall U.S. response distribution as an aggregate-population target and also evaluates 60 demographic groups, including group-conditioned steering and subgroup distribution matching. \\
\cite{yuan2024culturalpalette} & Country and Region & \Spl{}: Country and Region; \Dpl{}: Country and Region & High & Cultural Palette conditions on cultural/country targets and, on GlobalOpinionQA, compares generated answer distributions with country-specific human survey distributions. \\
\cite{zhao2024worldvaluesbench} & Subgroup & \Spl{}: Subgroup; \Dpl{}: Subgroup & High & WorldValuesBench evaluates demographic-group conditions, including intersectional groups, and compares model answer distributions with human answer distributions for those groups; country or region is not a standalone empirical target in the benchmark evaluation. \\
\cite{cao2025survey} & Country and Region & \Dpl{}: Country and Region & High & The method is trained and evaluated against country-level survey response distributions, making the country-level distribution the empirical target. \\
\cite{chen2025spica} & Subgroup & \Spl{}: Subgroup & High & SPICA retrieves evidence for explicit demographic groups and evaluates whether model behavior aligns with those subgroup conditions. \\
\cite{halpern2025calibrated} & Aggregate population & \Dpl{}: Aggregate population & High & The method calibrates a pooled population-level distribution of preferences without an explicit geographic or social partition. \\
\cite{luo2025sparse} & Country and Region & \Dpl{}: Country and Region & High & Its distributional evaluation uses GlobalOpinionQA country-level opinion distributions; perspective labels used for steering do not constitute empirically evaluated subgroup targets. \\
\cite{meister2025benchmarking} & Subgroup & \Spl{}: Subgroup; \Dpl{}: Subgroup & High & The benchmark explicitly evaluates whether models match the opinion distribution of a specified demographic group and whether group-conditioned steering improves that match. \\
\cite{suh2025subpop} & Subgroup & \Dpl{}: Subgroup & High & SubPOP contains subpopulation–response-distribution pairs and evaluates predictions against empirical response distributions for held-out subpopulations. \\
\cite{xu2025culturespa} & Country and Region & \Spl{}: Country and Region & High & CultureSPA explicitly steers models toward country/cultural targets and evaluates culture-specific alignment; it does not evaluate full response-frequency matching. \\
\cite{agarwal2026plural} & Country and Region & \Spl{}: Country and Region & High & PLURAL evaluates fidelity to target countries’ cultural/value profiles; it does not require matching a full empirical response-frequency distribution. \\
\cite{ali2026operationalizing} & Subgroup & \Spl{}: Subgroup & High & The study trains on and compares group-specific preferences and behavior, supporting subgroup steering; it does not evaluate subgroup response-distribution matching. \\
\cite{borah2026pact} & Country and Region & \Spl{}: Country and Region; \Dpl{}: Country and Region & High & PACT conditions on country/persona context and compares model response rates with human rates across country-level contexts. Demographic attributes are persona inputs, but subgroup-specific distributions are not the primary empirical target. \\
\cite{fu2026opgrpo} & Aggregate population & \Opl{}: Aggregate population & High & The method optimizes coverage of human-supported perspectives, with human perspective support defining the aggregate reference set. \\
\cite{kambhatla2026distributional} & Subgroup & \Dpl{}: Subgroup & High & The evaluation explicitly compares model-generated response distributions with ground-truth distributions for distinct population groups. \\
\cite{mundada2026pluraleval} & Aggregate population & \Opl{}: Aggregate population & High & PLURALEVAL derives a human-supported claim space from crowd discussions and evaluates model coverage of that aggregate viewpoint set. \\
\cite{oh2026scpo} & Country and Region & \Spl{}: Country and Region & High & SCPO explicitly optimizes and evaluates country/culture-conditioned preference fidelity rather than population response-frequency matching. \\
\cite{pooledayan2026overton} & Aggregate population & \Opl{}: Aggregate population & High & OvertonBench evaluates coverage of human-supported viewpoints elicited from an aggregate human study population. \\
\cite{russo2026moralgap} & Aggregate population & \Dpl{}: Aggregate population & High & The benchmark pairs moral dilemmas with aggregate distributions of human judgments and compares model and human judgment distributions. \\
\cite{tan2026subgroup} & Subgroup & \Spl{}: Subgroup & High & The study evaluates whether persona prompting reproduces subgroup-specific modal/value preferences; the target is subgroup-conditioned fidelity rather than the full subgroup response distribution. \\
\cite{yang2026coverage} & Aggregate population & \Opl{}: Aggregate population & Borderline & Borderline objective label: the authors frame the task as \Dpl, but the survey codes it as \Opl{} because the reported targets are coverage of the human response space and in-boundary validity rather than frequency matching. \\
\cite{zhu2026dvmap} & Subgroup & \Spl{}: Subgroup & High & DVMap conditions on high-consensus demographic groups and evaluates group-preference prediction. It filters substantial within-group disagreement, so it supports subgroup steering rather than subgroup distribution matching. \\
\cite{feng2024modular} & Country and Region; Subgroup & \Spl{}: Subgroup; \Dpl{}: Country and Region & High & Modular Pluralism evaluates demographic-conditioned steering on OpinionQA and compares generated probabilities with country-specific human survey distributions on GlobalOpinionQA. \\
\cite{shetty2025vital} & Country and Region; Subgroup & \Spl{}: Subgroup; \Dpl{}: Country and Region & High & VITAL evaluates steering toward explicit demographic groups and distributional matching on country-linked survey questions, yielding subgroup steering and country-level distributional assignments. \\
\cite{yao2025gdpo} & Country and Region & \Dpl{}: Country and Region & High & GDPO evaluates controllable opinion generation against country-specific survey distributions, with each country’s opinion distribution serving as the empirical target. \\
\cite{wu-etal-2026-demographic} & Country and Region & \Dpl{}: Country and Region & High & Country-level opinion distributions are the evaluation target; demographic cells are intermediate representations used to generate and aggregate perspectives. \\
\cite{chae2026cultural} & Country and Region & \Dpl{}: Country and Region & High & The evaluation explicitly measures within-country survey-distribution fidelity and preservation of cross-country divergence. \\
\cite{kotalawala2026plurva} & Country and Region & \Spl{}: Country and Region & High & The system receives a target country and is evaluated for country-consistent value judgments, supporting country-level steering. \\
\end{longtable}
\normalsize

\section{Full Direct/Adjacent Corpus Coding}\label{app:direct-manifest}
The following two companion tables provide the complete coding for the broader direct/adjacent corpus. Part I reports bibliographic, corpus-status, contribution, and venue information; Part II reports inclusion rationale, pluralistic objective, and objective-by-human-reference matrix assignment. The \emph{Reference} column uses the paper's numbered citation in the bibliography, which also links the two views of each paper.

\scriptsize
\setlength{\tabcolsep}{3pt}
\renewcommand{\arraystretch}{1.08}
\begin{longtable}{@{}L{0.50in}L{0.33in}L{2.55in}L{1.10in}L{0.82in}L{0.63in}@{}}
\caption{Direct/adjacent corpus coding, Part I: bibliographic, corpus-status, contribution, and venue information.}\label{tab:direct-manifest-a}\\
\toprule
\textbf{Ref.} & \textbf{Year} & \textbf{Title} & \textbf{Corpus status} & \textbf{Type} & \textbf{Venue} \\
\midrule
\endfirsthead
\multicolumn{6}{l}{\tablename\ \thetable\ (continued)}\\
\toprule
\textbf{Ref.} & \textbf{Year} & \textbf{Title} & \textbf{Corpus status} & \textbf{Type} & \textbf{Venue} \\
\midrule
\endhead
\midrule
\multicolumn{6}{r}{Continued on next page}\\
\endfoot
\bottomrule
\endlastfoot
\cite{durmus2023globalopinions} & 2023 & Towards Measuring the Representation of Subjective Global Opinions in Language Models & matrix-\allowbreak{}eligible socially grounded & Dataset/\allowbreak{}Benchmark & preprint \\
\cite{santurkar2023whose} & 2023 & Whose Opinions Do Language Models Reflect? & matrix-\allowbreak{}eligible socially grounded & Dataset/\allowbreak{}Benchmark & peer-reviewed \\
\cite{kirk2024prism} & 2024 & The PRISM Alignment Dataset: What Participatory, Representative and Individualised Human Feedback Reveals About the Subjective and Multicultural Alignment of Large Language Models & direct/\allowbreak{}adjacent outside matrix & Dataset/\allowbreak{}Benchmark & peer-reviewed \\
\cite{yuan2024culturalpalette} & 2024 & Cultural Palette: Pluralising Culture Alignment via Multi-agent Palette & matrix-\allowbreak{}eligible socially grounded & Method & preprint \\
\cite{zhao2024worldvaluesbench} & 2024 & WorldValuesBench: A Large-Scale Benchmark Dataset for Multi-Cultural Value Awareness of Language Models & matrix-\allowbreak{}eligible socially grounded & Dataset/\allowbreak{}Benchmark & peer-reviewed \\
\cite{cao2025survey} & 2025 & Specializing Large Language Models to Simulate Survey Response Distributions for Global Populations & matrix-\allowbreak{}eligible socially grounded & Method & peer-reviewed \\
\cite{castricato2025persona} & 2025 & PERSONA: A Reproducible Testbed for Pluralistic Alignment & direct/\allowbreak{}adjacent outside matrix & Dataset/\allowbreak{}Benchmark & peer-reviewed \\
\cite{chen2025spica} & 2025 & SPICA: Retrieving Scenarios for Pluralistic In-Context Alignment & matrix-\allowbreak{}eligible socially grounded & Method & peer-reviewed \\
\cite{halpern2025calibrated} & 2025 & Pairwise Calibrated Rewards for Pluralistic Alignment & matrix-\allowbreak{}eligible socially grounded & Method & preprint \\
\cite{lake2025distributional} & 2025 & From Distributional to Overton Pluralism: Investigating Large Language Model Alignment & direct/\allowbreak{}adjacent outside matrix & Empirical Analysis & peer-reviewed \\
\cite{luo2025sparse} & 2025 & Towards Low-Resource Alignment to Diverse Perspectives with Sparse Feedback & matrix-\allowbreak{}eligible socially grounded & Method & peer-reviewed \\
\cite{meister2025benchmarking} & 2025 & Benchmarking Distributional Alignment of Large Language Models & matrix-\allowbreak{}eligible socially grounded & Dataset/\allowbreak{}Benchmark & peer-reviewed \\
\cite{suh2025subpop} & 2025 & Language Model Fine-Tuning on Scaled Survey Data for Predicting Distributions of Public Opinions & matrix-\allowbreak{}eligible socially grounded & Method & peer-reviewed \\
\cite{xu2025culturespa} & 2025 & Self-Pluralising Culture Alignment for Large Language Models & matrix-\allowbreak{}eligible socially grounded & Method & peer-reviewed \\
\cite{agarwal2026plural} & 2026 & PLURAL: A Global Dataset for Value Alignment & matrix-\allowbreak{}eligible socially grounded & Dataset/\allowbreak{}Benchmark & preprint \\
\cite{ali2026operationalizing} & 2026 & Operationalizing Pluralistic Values in Large Language Model Alignment Reveals Trade-offs in Safety, Inclusivity, and Model Behavior & matrix-\allowbreak{}eligible socially grounded & Empirical Analysis & peer-reviewed \\
\cite{borah2026pact} & 2026 & Whose Norms? Disentangling Cultural and Personal Alignment in Large Language Models & matrix-\allowbreak{}eligible socially grounded & Empirical Analysis & preprint \\
\cite{freedman2026adaptive} & 2026 & Adaptive Pluralistic Alignment: A Pipeline for Dynamic Artificial Democracy & direct/\allowbreak{}adjacent outside matrix & Method & preprint \\
\cite{fu2026opgrpo} & 2026 & Overton Pluralistic Reinforcement Learning for Large Language Models & matrix-\allowbreak{}eligible socially grounded & Method & preprint \\
\cite{kambhatla2026distributional} & 2026 & Improving the Distributional Alignment of LLM & matrix-\allowbreak{}eligible socially grounded & Method & peer-reviewed \\
\cite{mundada2026pluraleval} & 2026 & Evaluating Language Model Pluralism through In-the-wild Crowd Discussions & matrix-\allowbreak{}eligible socially grounded & Dataset/\allowbreak{}Benchmark & peer-reviewed \\
\cite{oh2026scpo} & 2026 & Steerable Cultural Preference Optimization of Reward Models & matrix-\allowbreak{}eligible socially grounded & Method & preprint \\
\cite{pooledayan2026overton} & 2026 & Benchmarking Overton Pluralism in LLMs & matrix-\allowbreak{}eligible socially grounded & Dataset/\allowbreak{}Benchmark & peer-reviewed \\
\cite{russo2026moralgap} & 2026 & The Pluralistic Moral Gap: Understanding Moral Judgment and Value Differences between Humans and Large Language Models & matrix-\allowbreak{}eligible socially grounded & Empirical Analysis & peer-reviewed \\
\cite{tan2026subgroup} & 2026 & Can Persona-Prompted LLMs Emulate Subgroup Values? An Empirical Analysis of Generalisability and Fairness in Cultural Alignment & matrix-\allowbreak{}eligible socially grounded & Empirical Analysis & peer-reviewed \\
\cite{yang2026coverage} & 2026 & Where Models Converge and Humans Diverge: A Coverage Framework for Distributional Pluralism in Open-Ended Generation & matrix-\allowbreak{}eligible socially grounded & Evaluation Framework & preprint \\
\cite{zhang2026community} & 2026 & Cultivating Pluralism In Algorithmic Monoculture: The Community Alignment Dataset & direct/\allowbreak{}adjacent outside matrix & Dataset/\allowbreak{}Benchmark & peer-reviewed \\
\cite{zhu2026dvmap} & 2026 & DVMap: Fine-Grained Pluralistic Value Alignment via High-Consensus Demographic-Value Mapping & matrix-\allowbreak{}eligible socially grounded & Method & peer-reviewed \\
\cite{hwang2023useropinions} & 2023 & Aligning Language Models to User Opinions & direct/\allowbreak{}adjacent outside matrix & Method & peer-reviewed \\
\cite{chen2024pal} & 2024 & PAL: Pluralistic Alignment Framework for Learning from Heterogeneous Preferences & direct/\allowbreak{}adjacent outside matrix & Method & peer-reviewed \\
\cite{feng2024modular} & 2024 & Modular Pluralism: Pluralistic Alignment via Multi-LLM Collaboration & matrix-\allowbreak{}eligible socially grounded & Method & peer-reviewed \\
\cite{siththaranjan2024dpl} & 2024 & Distributional Preference Learning: Understanding and Accounting for Hidden Context in RLHF & direct/\allowbreak{}adjacent outside matrix & Method & peer-reviewed \\
\cite{sorensen2024kaleidoscope} & 2024 & Value Kaleidoscope: Engaging AI with Pluralistic Human Values, Rights, and Duties & direct/\allowbreak{}adjacent outside matrix & Dataset/\allowbreak{}Framework & peer-reviewed \\
\cite{sorensen2024roadmap} & 2024 & Position: A Roadmap to Pluralistic Alignment & direct/\allowbreak{}adjacent outside matrix & Position/\allowbreak{}Conceptual & peer-reviewed \\
\cite{adams2025steerable} & 2025 & Steerable Pluralism: Pluralistic Alignment via Few-Shot Comparative Regression & direct/\allowbreak{}adjacent outside matrix & Method/\allowbreak{}Benchmark & peer-reviewed \\
\cite{jiang2025individualistic} & 2025 & Can Language Models Reason about Individualistic Human Values and Preferences? & direct/\allowbreak{}adjacent outside matrix & Benchmark/\allowbreak{}Analysis & peer-reviewed \\
\cite{kumar2025compo} & 2025 & ComPO: Community Preferences for Language Model Personalization & direct/\allowbreak{}adjacent outside matrix & Method & peer-reviewed \\
\cite{shetty2025vital} & 2025 & VITAL: A New Dataset for Benchmarking Pluralistic Alignment in Healthcare & matrix-\allowbreak{}eligible socially grounded & Dataset/\allowbreak{}Benchmark & peer-reviewed \\
\cite{sorensen2025spectrum} & 2025 & Spectrum Tuning: Post-Training for Distributional Coverage and In-Context Steerability & direct/\allowbreak{}adjacent outside matrix & Method/\allowbreak{}Benchmark & preprint \\
\cite{sorensen2025valueprofiles} & 2025 & Value Profiles for Encoding Human Variation & direct/\allowbreak{}adjacent outside matrix & Method/\allowbreak{}Representation & peer-reviewed \\
\cite{srewa2025plurallm} & 2025 & PluralLLM & direct/\allowbreak{}adjacent outside matrix & Method & preprint \\
\cite{yao2025gdpo} & 2025 & No Preference Left Behind: Group Distributional Preference Optimization & matrix-\allowbreak{}eligible socially grounded & Method & peer-reviewed \\
\cite{zhang2025cot} & 2025 & Exploring Chain-of-Thought Reasoning for Steerable Pluralistic Alignment & direct/\allowbreak{}adjacent outside matrix & Method & peer-reviewed \\
\cite{zhong2025healthcare} & 2025 & Pluralistic Alignment for Healthcare: A Role-Driven Framework & direct/\allowbreak{}adjacent outside matrix & Method & peer-reviewed \\
\cite{imai2026mixdpo} & 2026 & MixDPO & direct/\allowbreak{}adjacent outside matrix & Method & preprint \\
\cite{kim2026valueflow} & 2026 & VALUEFLOW: Toward Pluralistic and Steerable Value-based Alignment in Large Language Models & direct/\allowbreak{}adjacent outside matrix & Method/\allowbreak{}Framework & preprint \\
\cite{lamerton2026eggroll} & 2026 & EGGROLL-IPO & direct/\allowbreak{}adjacent outside matrix & Method & peer-reviewed \\
\cite{mahbub2026coherence} & 2026 & Coherence Maximization Improves Pluralistic Alignment & direct/\allowbreak{}adjacent outside matrix & Method & preprint \\
\cite{pooledayan2026impact} & 2026 & A Roadmap to Impactful Pluralistic Alignment Research & direct/\allowbreak{}adjacent outside matrix & Position/\allowbreak{}Conceptual & preprint \\
\cite{srewa2026appa} & 2026 & APPA & direct/\allowbreak{}adjacent outside matrix & Method & preprint \\
\cite{vishwarupe2026repair} & 2026 & From Sycophantic Consensus to Pluralistic Repair: Why AI & direct/\allowbreak{}adjacent outside matrix & Framework & preprint \\
\cite{wang2026vrolora} & 2026 & V-RoLoRA: RLVR-Driven MoE Routing for Steerable Pluralistic Alignment & direct/\allowbreak{}adjacent outside matrix & Method & peer-reviewed \\
\cite{zhao2026lens} & 2026 & Choose Your Lens: Multi-Perspective Value Alignment of Chain-of-Thought Reasoning & direct/\allowbreak{}adjacent outside matrix & Method & peer-reviewed \\
\cite{zheng2026vispa} & 2026 & VISPA: Pluralistic Alignment via Automatic Value Selection and Activation & direct/\allowbreak{}adjacent outside matrix & Method & preprint \\
\cite{pan2026fgdalign} & 2026 & FGD-Align: Pluralistic Alignment for Large Language Models via Fuzzy Group Decision-Making & direct/\allowbreak{}adjacent outside matrix & Method & peer-reviewed \\
\cite{bhattacharyya2026alpha} & 2026 & ALPHA: Action-Based Learning for Pluralistic Human Alignment in Large Language Models & direct/\allowbreak{}adjacent outside matrix & Method & peer-reviewed \\
\cite{zhou2026disalign} & 2026 & Disentangling Consensus and Value-Specific Representations for Controllable Pluralistic Value Alignment of LLMs & direct/\allowbreak{}adjacent outside matrix & Method & peer-reviewed \\
\cite{naseem2026demosafety} & 2026 & Do Large Language Models Reflect Demographic Pluralism in Safety? & direct/\allowbreak{}adjacent outside matrix & Dataset/\allowbreak{}Benchmark & peer-reviewed \\
\cite{engelmann2025compatibility} & 2025 & LLMs' Pluralistic Compatibility & direct/\allowbreak{}adjacent outside matrix & Position/\allowbreak{}Conceptual & peer-reviewed \\
\cite{jiang2026picaco} & 2026 & PICACO: Pluralistic In-Context Value Alignment of LLMs via Total Correlation Optimization & direct/\allowbreak{}adjacent outside matrix & Method & peer-reviewed \\
\cite{guo2025couple} & 2025 & Counterfactual Reasoning for Steerable Pluralistic Value Alignment of Large Language Models & direct/\allowbreak{}adjacent outside matrix & Method & peer-reviewed \\
\cite{xu2026rushes} & 2026 & Rushes: A Human Preference Dataset for Pluralistic Alignment & direct/\allowbreak{}adjacent outside matrix & Dataset/\allowbreak{}Benchmark & preprint \\
\cite{tehenan2025mpta} & 2025 & MPTA: MultiTask Personalization Assessment & direct/\allowbreak{}adjacent outside matrix & Dataset/\allowbreak{}Benchmark & peer-reviewed \\
\cite{agarwal2026geometry} & 2026 & Geometry of Values: Task Vector Composition for Ethical Preference Alignment in Language Models & direct/\allowbreak{}adjacent outside matrix & Method/\allowbreak{}Benchmark & workshop \\
\cite{tan2026beyondcompromise} & 2026 & Beyond Compromise: Pareto-Lenient Consensus for Efficient Multi-Preference LLM Alignment & direct/\allowbreak{}adjacent outside matrix & Method & preprint \\
\cite{wu-etal-2026-demographic} & 2026 & Demographic Pluralism: Inference-Time Modeling of Pluralistic Human Preference Distributions & matrix-\allowbreak{}eligible socially grounded & Method & preprint \\
\cite{chae2026cultural} & 2026 & Cultural Divergence Preservation: Diagnosing Flattening and Caricature in LLM-Simulated Survey Populations & matrix-\allowbreak{}eligible socially grounded & Evaluation Framework & workshop \\
\cite{kotalawala2026plurva} & 2026 & PlurVA-LLM-2026 Shared Task Track-1: Pluralistic Value Alignment in LLMs via Multilingual Fine-Tuning and Threshold Calibration & matrix-\allowbreak{}eligible socially grounded & Method/\allowbreak{}System & workshop \\
\cite{tsoi2026dial} & 2026 & Which Objectives Need a Dial? Predicting Objective Conflict and Covering Trade-offs in Steerable Pluralistic Alignment & direct/\allowbreak{}adjacent outside matrix & Empirical Analysis & preprint \\
\cite{mohan2026heterogeneous} & 2026 & Learning Heterogeneous Preferences & direct/\allowbreak{}adjacent outside matrix & Method & preprint \\
\end{longtable}
\normalsize

\scriptsize
\setlength{\tabcolsep}{3pt}
\renewcommand{\arraystretch}{1.08}
\begin{longtable}{@{}L{0.52in}L{2.10in}L{1.70in}L{1.68in}@{}}
\caption{Direct/adjacent corpus coding, Part II: inclusion rationale, objective, and matrix assignment.}\label{tab:direct-manifest-b}\\
\toprule
\textbf{Ref.} & \textbf{Reason/reference type} & \textbf{Pluralistic or adjacent objective} & \textbf{Matrix assignments} \\
\midrule
\endfirsthead
\multicolumn{4}{l}{\tablename\ \thetable\ (continued)}\\
\toprule
\textbf{Ref.} & \textbf{Reason/reference type} & \textbf{Pluralistic or adjacent objective} & \textbf{Matrix assignments} \\
\midrule
\endhead
\midrule
\multicolumn{4}{r}{Continued on next page}\\
\endfoot
\bottomrule
\endlastfoot
\cite{durmus2023globalopinions} & explicit human reference & \Dpl; \Spl & \Spl{}: Country and Region; \Dpl{}: Country and Region \\
\cite{santurkar2023whose} & explicit human reference & \Dpl; \Spl & \Spl{}: Subgroup; \Dpl{}: Aggregate population; \Dpl{}: Subgroup \\
\cite{kirk2024prism} & Socially grounded evidence, pluralistic objective unspecified & Participatory/individualized preference feedback; objective target unspecified &  \\
\cite{yuan2024culturalpalette} & explicit human reference & \Dpl; \Spl & \Spl{}: Country and Region; \Dpl{}: Country and Region \\
\cite{zhao2024worldvaluesbench} & explicit human reference & \Dpl; \Spl & \Spl{}: Subgroup; \Dpl{}: Subgroup \\
\cite{cao2025survey} & explicit human reference & \Dpl & \Dpl{}: Country and Region \\
\cite{castricato2025persona} & Community, role, or persona representation & \Spl &  \\
\cite{chen2025spica} & explicit human reference & \Spl & \Spl{}: Subgroup \\
\cite{halpern2025calibrated} & explicit human reference & \Dpl & \Dpl{}: Aggregate population \\
\cite{lake2025distributional} & Objective/perspective defined; human reference unspecified & \Opl; \Dpl &  \\
\cite{luo2025sparse} & explicit human reference & \Dpl & \Dpl{}: Country and Region \\
\cite{meister2025benchmarking} & explicit human reference & \Dpl; \Spl & \Spl{}: Subgroup; \Dpl{}: Subgroup \\
\cite{suh2025subpop} & explicit human reference & \Dpl & \Dpl{}: Subgroup \\
\cite{xu2025culturespa} & explicit human reference & \Spl & \Spl{}: Country and Region \\
\cite{agarwal2026plural} & explicit human reference & \Spl & \Spl{}: Country and Region \\
\cite{ali2026operationalizing} & explicit human reference & \Spl & \Spl{}: Subgroup \\
\cite{borah2026pact} & explicit human reference & \Dpl; \Spl & \Spl{}: Country and Region; \Dpl{}: Country and Region \\
\cite{freedman2026adaptive} & Adjacent normative objective outside the three-objective taxonomy & Adjacent aggregation/social choice &  \\
\cite{fu2026opgrpo} & explicit human reference & \Opl & \Opl{}: Aggregate population \\
\cite{kambhatla2026distributional} & explicit human reference & \Dpl & \Dpl{}: Subgroup \\
\cite{mundada2026pluraleval} & explicit human reference & \Opl & \Opl{}: Aggregate population \\
\cite{oh2026scpo} & explicit human reference & \Spl & \Spl{}: Country and Region \\
\cite{pooledayan2026overton} & explicit human reference & \Opl & \Opl{}: Aggregate population \\
\cite{russo2026moralgap} & explicit human reference & \Dpl & \Dpl{}: Aggregate population \\
\cite{tan2026subgroup} & explicit human reference & \Spl & \Spl{}: Subgroup \\
\cite{yang2026coverage} & explicit human reference & \Opl & \Opl{}: Aggregate population \\
\cite{zhang2026community} & Socially grounded evidence, pluralistic objective unspecified & Pluralistic preference dataset; objective target unspecified &  \\
\cite{zhu2026dvmap} & explicit human reference & \Spl & \Spl{}: Subgroup \\
\cite{hwang2023useropinions} & Individual / personalization boundary & \Spl &  \\
\cite{chen2024pal} & Latent or method-defined preference structure & \Dpl &  \\
\cite{feng2024modular} & explicit human reference & \Opl; \Spl; \Dpl & \Spl{}: Subgroup; \Dpl{}: Country and Region \\
\cite{siththaranjan2024dpl} & Latent or method-defined preference structure & \Dpl &  \\
\cite{sorensen2024kaleidoscope} & Objective/perspective defined; human reference unspecified & \Opl &  \\
\cite{sorensen2024roadmap} & Objective/perspective defined; human reference unspecified & \Opl; \Spl; \Dpl &  \\
\cite{adams2025steerable} & Individual / personalization boundary & \Spl &  \\
\cite{jiang2025individualistic} & Individual / personalization boundary & \Spl &  \\
\cite{kumar2025compo} & Community, role, or persona representation & \Spl &  \\
\cite{shetty2025vital} & explicit human reference & \Opl; \Spl; \Dpl & \Spl{}: Subgroup; \Dpl{}: Country and Region \\
\cite{sorensen2025spectrum} & Objective/perspective defined; human reference unspecified & Steerable; Distributional &  \\
\cite{sorensen2025valueprofiles} & Individual / personalization boundary & \Spl &  \\
\cite{srewa2025plurallm} & Latent or method-defined preference structure & Unspecified heterogeneous-preference alignment &  \\
\cite{yao2025gdpo} & explicit human reference & \Dpl & \Dpl{}: Country and Region \\
\cite{zhang2025cot} & Objective/perspective defined; human reference unspecified & \Spl &  \\
\cite{zhong2025healthcare} & Community, role, or persona representation & \Opl; \Spl; \Dpl &  \\
\cite{imai2026mixdpo} & Latent or method-defined preference structure & Distributional-adjacent &  \\
\cite{kim2026valueflow} & Objective/perspective defined; human reference unspecified & \Spl &  \\
\cite{lamerton2026eggroll} & Adjacent normative objective outside the three-objective taxonomy & Adjacent social choice &  \\
\cite{mahbub2026coherence} & Community, role, or persona representation & \Spl &  \\
\cite{pooledayan2026impact} & Objective/perspective defined; human reference unspecified & \Opl; \Spl; \Dpl &  \\
\cite{srewa2026appa} & Adjacent normative objective outside the three-objective taxonomy & Adjacent aggregation/fairness &  \\
\cite{vishwarupe2026repair} & Adjacent normative objective outside the three-objective taxonomy & Adjacent interaction/process &  \\
\cite{wang2026vrolora} & Objective/perspective defined; human reference unspecified & \Spl &  \\
\cite{zhao2026lens} & Objective/perspective defined; human reference unspecified & \Spl &  \\
\cite{zheng2026vispa} & Objective/perspective defined; human reference unspecified & \Opl; \Spl; \Dpl &  \\
\cite{pan2026fgdalign} & Latent or method-defined preference structure & Pluralistic preference aggregation &  \\
\cite{bhattacharyya2026alpha} & Socially grounded evidence, pluralistic objective unspecified & Pluralistic human alignment; objective unspecified &  \\
\cite{zhou2026disalign} & Objective/perspective defined; human reference unspecified & Steerable value pluralism &  \\
\cite{naseem2026demosafety} & Socially grounded evidence, pluralistic objective unspecified & Pluralistic safety evaluation; objective unspecified &  \\
\cite{engelmann2025compatibility} & Objective/perspective defined; human reference unspecified & Conceptual/procedural and behavioral pluralism &  \\
\cite{jiang2026picaco} & Objective/perspective defined; human reference unspecified & Steerable value pluralism &  \\
\cite{guo2025couple} & Objective/perspective defined; human reference unspecified & Steerable value pluralism &  \\
\cite{xu2026rushes} & Individual / personalization boundary & Individual preference benchmark / pluralism-personalization boundary &  \\
\cite{tehenan2025mpta} & Community, role, or persona representation & Pluralistic/persona-conditioned evaluation &  \\
\cite{agarwal2026geometry} & Objective/perspective defined, human reference unspecified & Steerable value pluralism &  \\
\cite{tan2026beyondcompromise} & Adjacent normative objective outside the three-objective taxonomy & Pareto / multi-preference consensus &  \\
\cite{wu-etal-2026-demographic} & explicit human reference & \Dpl & \Dpl{}: Country and Region \\
\cite{chae2026cultural} & explicit human reference & \Dpl & \Dpl{}: Country and Region \\
\cite{kotalawala2026plurva} & explicit human reference & \Spl & \Spl{}: Country and Region \\
\cite{tsoi2026dial} & Objective/perspective defined; human reference unspecified & \Spl &  \\
\cite{mohan2026heterogeneous} & Individual / personalization boundary & Heterogeneous preference learning / individuated utility modeling &  \\
\end{longtable}
\normalsize

\clearpage

\normalsize
\clearpage
\section{Work Beyond the Objective-by-Human-Reference Matrix}\label{app:outside}
\begin{table}[!htbp]
\centering
\scriptsize
\caption{Direct/adjacent work intentionally retained outside the objective-by-human-reference matrix. These papers are not ``missing'' from Table~\ref{tab:joint-taxonomy}; their evaluated reference or objective falls outside the narrowly defined matrix. Communities and roles can be social; their exclusion here is an operational coding choice.}
\label{tab:outside-matrix}
\setlength{\tabcolsep}{3pt}
\renewcommand{\arraystretch}{1.08}
\begin{tabularx}{\textwidth}{@{}>{\bfseries\raggedright\arraybackslash}p{0.22\textwidth} >{\raggedright\arraybackslash}X@{}}
\toprule
Reason outside 3-by-3 matrix & Direct/adjacent works \\
\midrule
Objective/perspective defined; human reference unspecified & \textcite{lake2025distributional}; \textcite{sorensen2024kaleidoscope}; \textcite{sorensen2024roadmap}; \textcite{sorensen2025spectrum}; \textcite{zhang2025cot}; \textcite{kim2026valueflow}; \textcite{pooledayan2026impact}; \textcite{wang2026vrolora}; \textcite{zhao2026lens}; \textcite{zheng2026vispa}; \textcite{zhou2026disalign}; \textcite{engelmann2025compatibility}; \textcite{jiang2026picaco}; \textcite{guo2025couple}; \textcite{agarwal2026geometry}; \textcite{tsoi2026dial} \\
Latent or method-defined preference structure & \textcite{chen2024pal}; \textcite{siththaranjan2024dpl}; \textcite{srewa2025plurallm}; \textcite{imai2026mixdpo}; \textcite{pan2026fgdalign} \\
Socially grounded evidence, pluralistic objective unspecified & \textcite{kirk2024prism}; \textcite{zhang2026community}; \textcite{bhattacharyya2026alpha}; \textcite{naseem2026demosafety} \\
Community, role, or persona representation & \textcite{castricato2025persona}; \textcite{kumar2025compo}; \textcite{zhong2025healthcare}; \textcite{mahbub2026coherence}; \textcite{tehenan2025mpta} \\
Individual / personalization boundary & \textcite{hwang2023useropinions}; \textcite{adams2025steerable}; \textcite{jiang2025individualistic}; \textcite{sorensen2025valueprofiles}; \textcite{xu2026rushes}; \textcite{mohan2026heterogeneous} \\
Adjacent normative objective outside the three-objective taxonomy & \textcite{freedman2026adaptive}; \textcite{lamerton2026eggroll}; \textcite{srewa2026appa}; \textcite{vishwarupe2026repair}; \textcite{tan2026beyondcompromise} \\
\bottomrule
\end{tabularx}
\end{table}

\section{Sensitivity of the Objective-by-Human-Reference Matrix}\label{app:sensitivity}
\begin{table}[!htbp]
\centering
\scriptsize
\caption{Sensitivity of the landscape counts to evidence maturity and coding judgment. ``Peer-reviewed only'' removes records whose version available by the coverage cutoff is an arXiv/preprint rather than an archival venue publication. ``High-confidence only'' removes Borderline coding assignments.}
\label{tab:sensitivity-counts}
\setlength{\tabcolsep}{4.3pt}
\begin{tabular}{lrrrrrrrr}
\toprule
Scope & Papers & Assign. & O & S & D & Pop. & Country & Subgrp. \\
\midrule
All socially grounded & 29 & 38 & 4 & 16 & 18 & 7 & 17 & 14 \\
Empirical only & 29 & 38 & 4 & 16 & 18 & 7 & 17 & 14 \\
Peer-reviewed only & 18 & 24 & 2 & 10 & 12 & 4 & 6 & 14 \\
High-confidence only & 28 & 37 & 3 & 16 & 18 & 6 & 17 & 14 \\
\bottomrule
\end{tabular}
\end{table}

\section{Additional Evaluation Considerations}\label{app:extended-evaluation}
\subsection{Perspectivist Evaluation}
Perspectivist evaluation asks whether models preserve annotator-level or response-distribution structure instead of scoring only against a majority label. Multi-annotator modeling can improve uncertainty estimates and preserve systematic disagreement \citep{davani2022disagreements}; response-distribution analysis provides a statistical view of variation across items \citep{homan2022annotator}. These ideas are directly useful for pluralistic evaluation when the target is not a single population distribution but a distribution over perspectives or annotator-conditioned outcomes.

Perspectivist evaluation is most valuable when disagreement has structure. If repeated judgments are available, researchers can estimate whether variation is primarily item-level ambiguity, stable annotator tendency, subgroup difference, or contextual change. This permits richer targets than a single empirical histogram. For example, a model can be evaluated on its ability to predict annotator-conditioned labels, reproduce pairwise disagreement rates, or preserve correlations among annotator responses across items.

This perspective also cautions against using one LLM judge as the final arbiter of pluralism. If human evaluators disagree systematically about validity or reasonableness, collapsing their judgments into one judge score recreates the majority-gold problem at the evaluation layer. Multi-annotator evaluation, reporting judge distributions, and retaining evaluator metadata can make second-order disagreement visible rather than hiding it.

Pluralistic alignment can therefore borrow both data structures and metrics from perspectivist NLP while maintaining a different behavioral target. The key transfer is methodological: preserve who disagrees with whom, on which items, and with what confidence before deciding how a system should act on that disagreement.

\subsection{Evaluation Across Human-Reference Levels}
Alignment at one granularity does not imply alignment at another. In particular,
\begin{equation}
P_M(y\mid x,c) \approx P_H(y\mid x,c)
\end{equation}
does not entail
\begin{equation}
P_M(y\mid x,c,d) \approx P_H(y\mid x,c,d).
\end{equation}
Country-level success may therefore hide subgroup-level failures. Evaluations should report macro-averages, worst-group errors, uncertainty intervals for small cells, and invariance tests where no empirical group difference exists.

Cross-granularity evaluation should explicitly compare aggregate and disaggregated performance. A model may match a national distribution because errors cancel across subgroups, even while every subgroup is miscalibrated; conversely, subgroup-conditioned estimates may be noisy even when the aggregate target is accurate. Reporting only one level makes it impossible to distinguish these cases. Macro-group divergence, worst-group divergence, and the gap between aggregate and disaggregated error should therefore accompany country- or population-level scores when subgroup labels are available.

Small cells require uncertainty-aware evaluation. Confidence intervals, bootstrap resampling, or hierarchical shrinkage can prevent unstable subgroup estimates from being interpreted as meaningful differences. Evaluations should also test invariance on items where subgroup distributions are statistically similar. This guards against models that exaggerate demographic distinctions and rewards methods for preserving similarity when the data support similarity.

A further issue is compositionality. Success for age and gender separately does not imply success for age-by-gender, and success for a national population does not imply success for a demographic subgroup within that country. Multi-attribute resources make it possible to test whether conditioning effects compose empirically or whether the model simply adds stereotyped signals. This is one of the clearest places where our human-reference axis produces testable evaluation requirements.

\subsection{Measurement Validity}
Pluralistic evaluation inherits a measurement problem: constructs such as culture, values, reasonableness, and preference are not self-defining. \citet{dev2026culture} explicitly separate conceptualization from operationalization and measurement for cultural intelligence. \citet{wu2025cultural} similarly show how benchmark conclusions change depending on which cultural group, element, and awareness scope is represented. These principles should be applied to pluralistic alignment: a metric is meaningful only relative to a justified target population and construct.

A pluralistic benchmark implicitly makes at least four measurement choices: the construct of interest (e.g., preference, value, reasonableness, cultural appropriateness), the population to which the construct refers, the observable used to operationalize it, and the metric used to compare model and human behavior. Weakness at any layer can invalidate an apparently precise score. For example, a country label may be a defensible sampling stratum but a weak operationalization of culture; a multiple-choice survey may measure response selection but not the full space of reasons behind the selection.

Construct validity should therefore be discussed alongside predictive performance. Dev et al.'s separation of conceptualization, operationalization, and measurement provides a useful template for cultural constructs \citep{dev2026culture}; the cultural-alignment benchmark survey similarly shows how conclusions depend on which groups and cultural elements are represented \citep{wu2025cultural}. For pluralistic alignment, benchmark authors should justify why the chosen human data constitute an appropriate reference for the claimed form of pluralism.

Validity also has temporal and contextual dimensions. Survey distributions can shift with events, wording, and data-collection context, while user preferences can change during interaction. A benchmark score should therefore be interpreted as alignment to a specified population under specified conditions, not as a timeless measure of ``human values.'' Making those boundary conditions explicit improves both scientific interpretation and deployment relevance.

\section{Extended Research Questions}\label{app:research}
\subsection{Representation Beyond Static Demographics}
Observed social groups are interpretable and useful for auditing, yet demographics can be weak proxies for individual preference. Communities, value profiles, latent clusters, and persona representations may predict behavior better, but they are harder to interpret socially and govern institutionally. The revised corpus makes this distinction explicit: a large body of direct pluralistic work sits outside the objective-by-human-reference matrix precisely because its reference object is value-, persona-, community-, latent-, or individual-defined rather than a named social population \citep{chen2024pal,kumar2025compo,sorensen2025valueprofiles,imai2026mixdpo,mahbub2026coherence}.

Hybrid models may offer a better compromise. A system can preserve socially meaningful labels for representational auditing while learning latent factors within each group to capture residual variation. The central questions are whether those latent dimensions remain stable across tasks and time, whether users can contest them, and when a temporary context-specific preference should update the representation. Dynamic approaches begin to make preference drift explicit \citep{freedman2026adaptive}, but longitudinal human data remain scarce.

\subsection{Aggregation, Preservation, and Normative Boundaries}
Representing heterogeneity does not determine what a system should do with it. Population weighting, equal weighting, consensus, max-min welfare, random social choice, user choice, and deliberative aggregation encode different normative priorities. Consensus-seeking \citep{bakker2022agreement}, collective constitutional aggregation \citep{huang2024collective}, max-min objectives \citep{chakraborty2024maxmin}, and newer social-choice approaches should therefore be treated as alternatives to distribution matching rather than silently folded into it.

Safety constraints further modify the aggregation problem. A pluralistic system need not support every expressed preference, but any boundary changes the viewpoint set relevant to \Opl, removes probability mass, or prevents certain steering targets. Researchers should therefore report what is constrained, who defines the constraint, and whether the rule suppresses some populations disproportionately. In high-stakes settings, both the aggregation rule and the safety boundary may require participatory or institutional justification rather than being selected only for predictive performance.

\subsection{Evaluation Beyond Single-Gold and Closed-Set Assumptions}
Pluralistic evaluation should not recreate a single-gold assumption at the meta-level. Human validation of perspective sets, sampling uncertainty in empirical distributions, and disagreement about evaluator criteria all complicate benchmarking. Perspective validity can be weighted probabilistically; group distributions can carry sampling intervals; and judge disagreement can itself be reported as a second-order distribution. Perspectivist NLP provides useful machinery for preserving rather than adjudicating away this uncertainty \citep{davani2022disagreements,homan2022annotator}.

Open-ended generation makes the problem harder because probability mass must be defined over a semantic response space rather than a fixed label set. Clustering can create brittle bins, embedding-space densities depend on the representation model, and LLM judges can import their own value priors. Promising approaches separate support coverage from in-boundary validity and make the semantic representation explicit \citep{yang2026coverage}. A mature framework will likely need repeated human responses, calibrated semantic neighborhoods, and uncertainty over the estimated response manifold.

\subsection{Deployment, Personalization, and Temporal Adaptation}
The field still has limited evidence that explicitly pluralistic objectives improve real-world outcomes. Deployment introduces choices that benchmarks usually hold fixed: which population is relevant, when a user may override defaults, how feedback updates the system, and how disagreement is communicated. Studies should therefore measure interaction-level outcomes such as discoverability of minority views, predictability of steering controls, perceived representation, trust, and decision quality rather than benchmark divergence alone \citep{pooledayan2026impact}.

Future assistants may also need to balance societal representation with user-specific utility. One architecture can maintain a population or group prior while updating a posterior from personal evidence, deciding when to personalize, when to expose alternatives, and when uncertainty is too high to infer a user preference. Evaluation should report both personalization regret and population-level representation so that gains for frequent users do not mask systematic loss of visibility for others. Temporal adaptation adds a further stability--responsiveness trade-off: systems that update too slowly become stale, while systems that update too quickly may overreact to transient behavior or manipulation.

\subsection{Multimodal Pluralism}
Multimodal pluralism remains underdeveloped. Key questions include whether culturally or socially conditioned judgments transfer from text to image-text inputs, whether visual cues induce unwanted identity inference, and how to construct human reference distributions for images, audio, and video. A natural extension of \Dpl{} is
\begin{equation}
P_M(y\mid x_{\text{text}},x_{\text{vision}},x_{\text{audio}},g) \approx P_H(y\mid x_{\text{text}},x_{\text{vision}},x_{\text{audio}},g),
\end{equation}
but subgroup-conditioned multimodal distributions require both diverse participants and controlled variation in modality. Paired designs that render the same underlying scenario in different modalities could identify whether a failure comes from the human-reference level or from the modality itself, and could support evaluation of accessibility preferences, culturally situated image interpretation, and disagreement over generated visual content.

\clearpage
\section{Extended Resource Inventory}\label{app:resources}
Pluralistic evaluation requires datasets that retain the structure of human disagreement. Tables in this appendix catalog the principal reusable resources in the surveyed corpus. To keep the resource map readable in portrait format, aggregate-population/country/subgroup/individual support is consolidated into a single \textbf{Supported granularity} column, while \textbf{Subgroup example} retains each resource's actual subgroup resolution.

\begin{table}[!htbp]
\centering
\scriptsize
\caption{Datasets and benchmark resources relevant to pluralistic alignment. \textbf{Resource} names the reusable dataset or benchmark; \textbf{Data source} records provenance; \textbf{Supported granularity} indicates whether the resource supports aggregate-population, country- and region-level, subgroup, or individual-adjacent analysis.}
\label{tab:datasets}
\setlength{\tabcolsep}{2.2pt}
\renewcommand{\arraystretch}{1.05}
\begin{tabularx}{\textwidth}{@{}>{\raggedright\arraybackslash}p{0.163\textwidth} >{\raggedright\arraybackslash}p{0.152\textwidth} >{\centering\arraybackslash}p{0.065\textwidth} >{\raggedright\arraybackslash}p{0.12\textwidth} >{\raggedright\arraybackslash}p{0.255\textwidth} X@{}}
\toprule
Resource & Data source & Mode & Supported granularity & Subgroup example & Primary use \\
\midrule
OpinionQA~\citep{santurkar2023whose} & Pew American Trends Panel & Text & Aggregate population; subgroup & 60 U.S. groups; e.g., age, gender, race/ethnicity, education, marital status, party/ideology & Demographic distributional evaluation and steering \\
GlobalOpinionQA \citep{durmus2023globalopinions} & World Values Survey + Pew Global Attitudes Survey & Text & Aggregate population; country & -- & Country-conditioned opinion distributions and steering \\
WorldValuesBench \citep{zhao2024worldvaluesbench} & World Values Survey respondent-level values + demographics & Text & Aggregate population; country; subgroup & 42 demographic variables in full data; probe groups use continent-by-urban/rural-by-four education levels & Demographic-conditioned value prediction and distributional awareness \\
SubPOP~\citep{suh2025subpop} & Pew American Trends Panel + General Social Survey & Text & Aggregate population; subgroup & 22 subpopulations across region, income, education, race/ethnicity, gender, religion, party, ideology & Fine-tuning to subpopulation response distributions \\
PRISM~\citep{kirk2024prism} & 1,500 recruited participants from 75 countries; live interactions with 21 LLMs & Text & Aggregate population; country; subgroup; individual-adjacent & Detailed profiles including age, gender, country/nationality, education, and other sociodemographics & Participatory and individualized preference feedback \\
VITAL~\citep{shetty2025vital} & Author-constructed healthcare value-laden situations and multiple-choice questions & Text & Aggregate population & -- & Domain-specific evaluation of all three pluralistic objectives in healthcare \\
\bottomrule
\end{tabularx}
\end{table}

\subsection{Aggregate-Population and \Opl{} Resources}
OvertonBench formalizes pluralistic coverage using human-grounded acceptable-perspective sets \citep{pooledayan2026overton}. PLURALEVAL and WildSCOPE extend evaluation to in-the-wild crowd discussions and open-ended claims \citep{mundada2026pluraleval}. VITAL transfers pluralistic evaluation to healthcare \citep{shetty2025vital}. These resources are important because lexical diversity or output entropy alone cannot establish that a model covers human-supported perspectives.

These resources differ in how the human reference set is constructed. OvertonBench uses explicit human judgments to define acceptable perspective coverage, whereas WildSCOPE/PLURALEVAL derives a claim space from naturally occurring crowd discussion; VITAL anchors disagreement in a specialized healthcare domain. The distinction matters because a benchmark can measure coverage only relative to the viewpoints that its collection process makes observable.

Population-level resources should therefore document both breadth and provenance. A benchmark sampled from one national platform may still be useful as a population-level target, but the relevant population must be named. For \Opl{} in particular, metadata about which people validated a perspective as reasonable is not incidental: it determines whose viewpoint window is being measured. This provides the foundation for future country- or subgroup-conditioned \Opl{} benchmarks.

\subsection{Country and Regional Resources}
GlobalOpinionQA is the canonical country-level opinion-distribution resource \citep{durmus2023globalopinions}. Survey-distribution specialization builds directly on global survey data to train models toward national distributions \citep{cao2025survey}. PRISM contains participants from many countries and retains detailed user profiles, allowing both aggregate and disaggregated analysis \citep{kirk2024prism}.

\begin{table}[!htbp]
\ContinuedFloat
\centering
\scriptsize
\caption{Datasets and benchmark resources relevant to pluralistic alignment (continued).}
\setlength{\tabcolsep}{2.2pt}
\renewcommand{\arraystretch}{1.05}
\begin{tabularx}{\textwidth}{@{}>{\raggedright\arraybackslash}p{0.163\textwidth} >{\raggedright\arraybackslash}p{0.152\textwidth} >{\centering\arraybackslash}p{0.065\textwidth} >{\raggedright\arraybackslash}p{0.12\textwidth} >{\raggedright\arraybackslash}p{0.255\textwidth} X@{}}
\toprule
Resource & Data source & Mode & Supported granularity & Subgroup example & Primary use \\
\midrule
ValuePrism / Value Kaleidoscope~\citep{sorensen2024kaleidoscope} & 31k human-written situations linked to 218k generated-and-human-validated values, rights, and duties & Text & Aggregate population & Diverse annotator backgrounds analyzed; released target is a pluralistic value set rather than subgroup distributions & Structured value-pluralism coverage and value reasoning \\
OvertonBench~\citep{pooledayan2026overton} & U.S.-representative Prolific study; value-laden political/social questions & Text & Aggregate population & Metadata include age, sex, ethnicity, political affiliation; target remains population-level & Human-grounded Overton coverage \\
WildSCOPE / PLURALEVAL~\citep{mundada2026pluraleval} & Natural in-the-wild crowd discussions and free-form responses & Text & Aggregate population & -- & Open-ended claim-space / Overton coverage \\
PERSONA / PERSONA Bench~\citep{castricato2025persona} & Procedurally generated profiles grounded in U.S. Census statistics + idiosyncratic attributes & Text & Persona; individual-adjacent & Multi-attribute synthetic personas: demographic + idiosyncratic features & Persona-based pluralistic evaluation / preference data \\
MPTA~\citep{tehenan2025mpta} & WVS + EVS + GSS survey records transformed into real hyper-granular personas & Text & Persona; subgroup-analysis; individual-adjacent & Demographics, beliefs, and values; subgroup-specific slices reveal hidden performance disparities & Persona-conditioned pluralistic alignment assessment \\
SPICA preference data~\citep{chen2025spica} & Human preference annotations on PRISM-derived scenarios & Text & Subgroup & Four U.S. groups: Democrat/Republican by regular religious participation (yes/no) & Group-aware in-context alignment and retrieval \\
DVMap data~\citep{zhu2026dvmap} & Demographic--value mappings used for fine-grained alignment & Text & Country; subgroup & Multi-dimensional demographic constraints rather than a single national label & Fine-grained demographic/value steering and optimization \\
\bottomrule
\end{tabularx}
\end{table}

Country resources also vary in whether they provide aggregate national frequencies, respondent-level records, or disaggregated profiles. GlobalOpinionQA emphasizes national response distributions; WorldValuesBench preserves richer World Values Survey demographics; PRISM combines country coverage with participant-level attributes. These formats support different research questions even when they draw from overlapping notions of national representation.

A useful benchmark-design principle is to retain respondent-level information whenever licensing and privacy permit, even if the headline task is country-level. Aggregated country frequencies support stable distributional evaluation, but respondent-level data allows researchers to test whether national success survives disaggregation. This avoids turning the original aggregation decision into a permanent ceiling on later subgroup analysis.

\subsection{Subgroup and Fine-Grained Resources}
OpinionQA and SubPOP are central for subgroup-level distributional work \citep{santurkar2023whose,suh2025subpop}. WorldValuesBench preserves respondent-level demographic information and, in its probe study, constructs multi-attribute groups from continent, urban/rural status, and education \citep{zhao2024worldvaluesbench}. DVMap contributes a high-consensus demographic-value corpus and explicitly evaluates cross-demographic generalization \citep{zhu2026dvmap}. PERSONA offers a controlled synthetic testbed that combines multiple attributes, while MPTA uses survey-grounded hyper-granular personas and shows that subgroup-specific evaluation can expose failures hidden by aggregate metrics \citep{castricato2025persona,tehenan2025mpta}. ALPHA adds behavior-derived demographic signals at much larger scale, while Demo-SafetyBench retains demographic metadata specifically to probe pluralism in safety judgments \citep{bhattacharyya2026alpha,naseem2026demosafety}. These resources expand the evidence base without automatically becoming objective-by-human-reference matrix papers: resource granularity and pluralistic objective are coded separately. The key limitation is that increasingly fine subgroup cells require much larger human samples to estimate stable distributions.

\begin{table}[!htbp]
\ContinuedFloat
\centering
\scriptsize
\caption{Datasets and benchmark resources relevant to pluralistic alignment (continued).}
\setlength{\tabcolsep}{2.2pt}
\renewcommand{\arraystretch}{1.05}
\begin{tabularx}{\textwidth}{@{}>{\raggedright\arraybackslash}p{0.163\textwidth} >{\raggedright\arraybackslash}p{0.152\textwidth} >{\centering\arraybackslash}p{0.065\textwidth} >{\raggedright\arraybackslash}p{0.12\textwidth} >{\raggedright\arraybackslash}p{0.255\textwidth} X@{}}
\toprule
Resource & Data source & Mode & Supported granularity & Subgroup example & Primary use \\
\midrule
ALPHA50M / OpinionQA-XL~\citep{bhattacharyya2026alpha} & Large-scale advertisement interactions with demographic engagement signals; broad opinion-survey evaluation & Text & Aggregate population; subgroup & Demographic engagement patterns provide behavior-derived group signals rather than only self-reported survey labels & Behavioral pluralistic alignment and broad opinion generalization \\
Demo-SafetyBench~\citep{naseem2026demosafety} & DICES-derived safety prompts retaining demographic metadata, expanded across 14 safety domains & Text & Subgroup & Demographic metadata are preserved to test variation in safety perception across communities & Demographic pluralism and safety-sensitivity evaluation \\
Singapore subgroup-value data~\citep{tan2026subgroup} & World Values Survey-derived structured subgroup preferences; $>$20k samples & Text & Country; subgroup & Fine-grained Singapore subgroups over attributes including age, gender, ethnicity, and religion & Subgroup value emulation, generalization, fairness \\
Community Alignment~\citep{zhang2026community} & Representative multilingual human preference study; nearly 200k comparisons across five countries & Text & Aggregate population; country; subgroup & Country samples include balancing/analysis across age, gender, ethnicity and related attributes & Learning heterogeneous preferences across national populations \\
PACT~\citep{borah2026pact} & 200-person five-country human study + culturally situated scenarios & Text & Aggregate population; country; subgroup & Analyses include country, age, gender; also separates personal choice from perceived norm & Cultural--personal trade-offs and human response distributions \\
PLURAL~\citep{agarwal2026plural} & Integrated Values Survey / WVS respondent records transformed into preference triplets & Text & Aggregate population; country; subgroup & Metadata include sex, age, education, region, and additional respondent demographics & Global value steering while preserving within-country diversity \\
Spectrum Suite~\citep{sorensen2025spectrum} & $>$40 source datasets unified into $>$90 conditional-distribution tasks & Text & Aggregate population; individual-adjacent & Mixes population, individual-preference, subjective-NLP, and synthetic distribution tasks rather than a fixed demographic hierarchy & Distributional coverage and in-context steerability \\
Value Profiles~\citep{sorensen2025valueprofiles} & Individual rating demonstrations compressed into natural-language value profiles & Text & Aggregate population; individual-adjacent & Latent/value-profile clusters can be compared with demographic groupings & Individual variation and simulated annotator populations \\
Rushes~\citep{xu2026rushes} & 44,226 sequential narrative choices from 8,167 users with persistent user identifiers & Text & Individual-adjacent & Repeated user trajectories rather than demographic cells & Sequential preference heterogeneity and majority-default diagnostics \\
Culture-centric VizWiz~\citep{karamolegkou2024vlm} & VizWiz images + survey-collected caption preferences from culturally diverse users & Vision + Text & Cultural; individual-adjacent & Cultural-background preferences; no fixed demographic hierarchy & Multimodal cultural preference evaluation \\
\bottomrule
\end{tabularx}
\end{table}

The subgroup resources in Table~\ref{tab:datasets} differ substantially in resolution and construction. OpinionQA exposes many single-axis demographic groups; SubPOP scales distribution prediction over a fixed set of subpopulations; WorldValuesBench retains dozens of respondent attributes; PACT and the Singapore subgroup-value data support multi-attribute analyses; PERSONA provides controlled synthetic combinations rather than directly observed human cells; and MPTA constructs real survey-grounded personas while using subgroup slices primarily for evaluation rather than defining a subgroup-conditioned alignment target. These differences mean that ``subgroup benchmark'' is not a single data regime.

Researchers should report which subgroup axes are observed, which are combined, the effective sample size of each target, and whether group labels are self-reported, inferred, or synthetically assigned. Fine-grained resources are especially vulnerable to unstable estimates when many attributes are crossed. Benchmark documentation should therefore include uncertainty or minimum-count rules, not just the list of available demographics.

\subsection{Individual-Adjacent and Sequential Resources}
Some resources retain repeated user-level behavior rather than only cross-sectional group labels. PRISM links detailed participant profiles to multiple conversations, Value Profiles compress repeated judgments into interpretable user-level representations, and Rushes records time-ordered choices from persistent users \citep{kirk2024prism,sorensen2025valueprofiles,xu2026rushes}. These resources are adjacent to pluralistic alignment because they expose heterogeneity that population averages hide, but their natural predictive target is an individual trajectory rather than a social distribution.

For pluralistic analysis, their value is in aggregation and decomposition. Repeated user records allow researchers to estimate within-user stability, between-user variation, and the amount of population error attributable to majority-default predictions. They also enable a stronger personalization boundary test: if an individual model improves user-level prediction, does aggregating those predictions recover the observed population distribution, or does the personalization mechanism systematically overfit frequent users? Sequential resources make this question dynamic by revealing preference persistence and drift rather than assuming independent static judgments.

\subsection{Benchmark Coverage and Cultural Scope}
Benchmark inventories themselves can be analyzed for representational coverage. \citet{wu2025cultural} review 105 cultural-alignment benchmarks and find large imbalances in Cultural Group, Cultural Elements, and Awareness Scope, including a 97.1\% majority-focused rate. For pluralistic alignment, this suggests that benchmark coverage should be audited not only by geography but also by whether datasets preserve internal diversity and whether the unit of aggregation matches the deployment claim.

Accordingly, resource documentation for pluralistic alignment should answer at least five questions: who was sampled, which attributes were retained, what unit was aggregated, how disagreement was encoded, and which outputs were available for models to choose or generate. These choices jointly determine the kinds of pluralistic claims the benchmark can support. A dataset may be globally broad yet internally majority-focused, or subgroup-rich yet limited to a narrow task domain.

Candidate-response support is another underappreciated form of benchmark bias. If human raters only compare model-generated candidates that already omit minority-supported responses, the resulting preference data cannot reveal those missing modes. Community Alignment highlights this issue at the preference-data level \citep{zhang2026community}. Resource audits should therefore consider not only participant diversity but also the diversity and provenance of the response space presented to those participants.

\subsection{Multimodal Resources}
Multimodal perspectivist and cultural work provides early building blocks rather than mature pluralistic benchmarks. \citet{viridiano2022perspective} argue for perspective-preserving annotation in multimodal datasets; \citet{karamolegkou2024vlm} introduce a culture-centric VLM evaluation using caption preferences; and \citet{yadav2025beyond} evaluate cultural value sensitivity in multimodal models. A major open opportunity is a benchmark that directly combines the three pluralistic objectives with country or subgroup-conditioned multimodal human distributions.

Multimodal resources add additional provenance dimensions: who produced the image or audio, who is depicted, who judges the output, and which modality carries the social context. A dataset can be culturally diverse in its images but homogeneous in its raters, or vice versa. These configurations answer different questions and should be reported separately.

The current gap is not simply a shortage of VLM benchmarks; it is a shortage of multimodal benchmarks with \emph{pluralistic human reference structure}. A mature resource would retain repeated human judgments, subgroup or country metadata where appropriate, and modality-controlled variants that test whether the target distribution changes when equivalent information is presented as text, image, or both. Such designs would allow evaluation for the three pluralistic objectives to move beyond text without reducing multimodal pluralism to generic cultural accuracy.

\end{document}